\documentclass[a4paper,fleqn]{cas-dc}
\usepackage{amsmath}
\usepackage{amssymb}
\usepackage{mathtools}
\usepackage{physics}
\usepackage[square, comma, sort&compress, numbers]{natbib}
\usepackage{algorithm}
\usepackage{algpseudocode}
\usepackage{subfigure}
\usepackage{epsfig}
\usepackage{multicol}
\usepackage{booktabs}
\usepackage{xcolor}
\usepackage{graphics}
\usepackage{multirow}
\usepackage{svg}
\usepackage{amsthm}
\usepackage{graphicx} 
\usepackage{pifont}
\usepackage{amsfonts}
\usepackage{array}
\usepackage{multicol}
\usepackage{cleveref}
\usepackage{colortbl}
\usepackage{changepage}
\usepackage{bm}
\usepackage[final, markup=none, commandnameprefix=always]{changes}
\newtheorem{Assumption}{Assumption}
\newtheorem{Theorem}{Theorem}
\newtheorem{Remark}{Remark}

\graphicspath{{images/}}
\graphicspath{{bios/}}

\makeatletter
\newcommand{\frameglobal}{\texttt{G}}
\newcommand{\framebody}{\texttt{B}}
\newcommand{\framewheel}{\texttt{W}}
\newcommand{\framewheelij}{\texttt{W}_{ij}}

\def\frG{\frameglobal}
\def\frB{\framebody}
\def\frW{\framewheel}
\def\frWij{\frW ij}

\newcommand{\globX}{x_\frG}
\newcommand{\globY}{y_\frG}

\newcommand{\globPhi}{\varphi_\frG}

\newcommand{\globVX}{\dot{x}_\frG}
\newcommand{\globVY}{\dot{y}_\frG}

\newcommand{\globVPhi}{\dot{\varphi}_\frG}

\newcommand{\vehicleVX}{\dot{x}_\frB}
\newcommand{\vehicleVY}{\dot{y}_\frB}

\newcommand{\vehicleVPhi}{\dot{\varphi}_\frB}

\newcommand{\globAX}{\ddot{x}_\frG}
\newcommand{\globAY}{\ddot{y}_\frG}

\newcommand{\globAPhi}{\ddot{\varphi}_\frG}

\newcommand{\vehicleAX}{\ddot{x}_\frB}
\newcommand{\vehicleAY}{\ddot{y}_\frB}

\newcommand{\vehiclAPhi}{\ddot{\varphi}_\frB}

\newcommand{\globO}{O_\frG}
\newcommand{\globXvec}{\vec{x}_\frG}
\newcommand{\globYvec}{\vec{y}_\frG}
\newcommand{\globZvec}{\vec{z}_\frG}

\newcommand{\vehicleO}{O_\frB}
\newcommand{\vehicleXvec}{\vec{x}_\frB}
\newcommand{\vehicleYvec}{\vec{y}_\frB}
\newcommand{\vehicleZvec}{\vec{z}_\frB}

\newcommand{\wheelOij}{O_{\frWij}}
\newcommand{\wheelXijvec}{\vec{x}_{\frWij}}
\newcommand{\wheelYijvec}{\vec{y}_{\frWij}}
\newcommand{\wheelZijvec}{\vec{z}_{\frWij}}

\newcommand{\vehicleYaw}{\globPhi}
\newcommand{\wheelyaw}{\delta_{ij}}

\newcommand{\FBxij}{\prescript{x}{B}{F_{ij}}}
\newcommand{\FByij}{\prescript{y}{B}{F_{ij}}}

\newcommand{\Fwxij}{\prescript{x}{W}{F_{ij}}}
\newcommand{\Fwyij}{\prescript{y}{W}{F_{ij}}}

\newcommand{\Vvij}{\prescript{}{B}{\bm{v}_{ij}}}
\newcommand{\Vvxij}{\prescript{x}{B}{v_{ij}}}
\newcommand{\Vvyij}{\prescript{y}{B}{v_{ij}}}

\newcommand{\Vwxij}{\prescript{x}{W}{v_{ij}}}
\newcommand{\Vwyij}{\prescript{y}{W}{v_{ij}}}

\newcommand{\omegawij}{\prescript{}{w}{\omega_{ij}}}
\newcommand{\omegasij}{\prescript{}{s}{\omega_{ij}}}

\newcommand{\Psij}{{P_{sij}}}
\newcommand{\Palphaij}{{P_{\alpha ij}}}
\newcommand{\Ptij}{{P_{tij}}}
\makeatother

\def\tsc#1{\csdef{#1}{\textsc{\lowercase{#1}}\xspace}}
\tsc{WGM}
\tsc{QE}
\usepackage{xcolor}
\usepackage{xpatch}
\makeatletter
\ExplSyntaxOn
\cs_new:Npn \bibColoredItems #1#2
{
	\clist_map_inline:nn {#2} { \cs_new:cpn {bib@colored@##1} {#1} }
}
\ExplSyntaxOff
\newcommand\bib@setcolor[1]{%
	\ifcsname bib@colored@#1\endcsname
	\expandafter\color\expandafter{\csname bib@colored@#1\endcsname}
	\else
	\normalcolor
	\fi
}

\xpatchcmd\@bibitem
{\item}
{\bib@setcolor{#1}\item}
{}{}

\xpatchcmd\@lbibitem
{\item}
{\bib@setcolor{#2}\item}
{}{}
\makeatother

\bibColoredItems{blue}{ }
\bibColoredItems{red}{ }

\begin{document}
	\let\WriteBookmarks\relax
	\def\floatpagepagefraction{1}
	\def\textpagefraction{.001}
	\shorttitle{}
	\shortauthors{}
	\title [mode = title]{Tire Wear Aware Trajectory Tracking Control for Multi-axle Swerve-drive Autonomous Mobile Robots}

\author{Tianxin Hu$^\dagger$}[orcid=0009-0009-0308-424X]
\ead{tianxin001@e.ntu.edu.sg}

\author{Xinhang Xu$^\dagger$}[orcid=0009-0000-4850-2278]
\ead{xinhang.xu@ntu.edu.sg}

\author{Thien-Minh Nguyen}[orcid=0000-0003-1315-0967]
\ead{thienminh.nguyen@ntu.edu.sg}

\author{Fen Liu}[orcid=0000-0002-3757-0838]
\ead{fen.liu@ntu.edu.sg}

\author{Shenghai Yuan*}[orcid=0009-0003-1887-6342]
\ead{shyuan@ntu.edu.sg}

\author{Lihua Xie}[orcid=0000-0002-7137-4136]
\ead{elxie@ntu.edu.sg}

\cortext[cor1]{Corresponding author, $^\dagger$ Equal Contribution.}

\affiliation{organization={Nanyang Technological University, Nanyang Avenue},  postcode={639798},   country={Singapore}}

\begin{abstract}
Multi-axle Swerve-drive Autonomous Mobile Robots (MS-AGVs) equipped with independently steerable wheels are commonly used for high-payload transportation. In this work, we present a novel model predictive control (MPC) method for MS-AGV trajectory tracking that takes tire wear minimization consideration in the objective function.
To speed up the problem-solving process, we propose a hierarchical controller design and simplify the dynamic model by integrating the \textit{magic formula tire model} and \textit{simplified tire wear model}. In the experiment, the proposed method can be solved by simulated annealing in real-time on a normal personal computer and by incorporating tire wear into the objective function, tire wear is reduced by 19.19\% while maintaining the tracking accuracy in curve-tracking experiments. In the more challenging scene: the desired trajectory is offset by 60 degrees from the vehicle's heading, the reduction in tire wear increased to 65.20\% compared to the kinematic model without considering the tire wear optimization.
\end{abstract}

\maketitle
	
\section{Introduction}
Multi-axle \cite{li2023integrated} Swerve-drive Autonomous Guided Vehicle (MS-AGV) is a type of heavy-duty vehicle equipped with multiple independently controlled steering wheels. This design provides MS-AGVs with a unique combination of high load capacity \cite{beyersdorfer2013novel,tang2024all} and exceptional maneuverability \cite{zhu2025local,gan2023dp}, making them highly suitable for complex industrial environments \cite{Bai2025Realm,ma2024mm}, such as automated warehouses and port logistics \cite{nguyen2022ntu,nguyen2024mcd,zhao2019modelling,islam2020planning,williams2012generalised}. 
However, effectively controlling MS-AGVs presents several challenges. These include achieving accurate kino-dynamic modeling \cite{10900461}, ensuring precise trajectory tracking \cite{cao2022direct}, and optimizing speed for operational efficiency \cite{xu2024cost,Nguyen2024GPTR}. Recent works have explored prescribed performance control under uncertainties and faults, such as \cite{zhang2024fault, zhang2022neural}, but they do not consider tire wear, which is critical in MS-AGV applications. Furthermore, practical concerns such as minimizing tire wear, which directly impacts maintenance costs, add complexity to the problem \cite{cao2023path}. Despite significant advancements, no existing solution \cite{wang2023multi} comprehensively addresses these issues in an integrated manner, leaving a critical gap in MS-AGV planning and control strategies.

\begin{figure}
    \centering
    \includegraphics[width=1\linewidth]{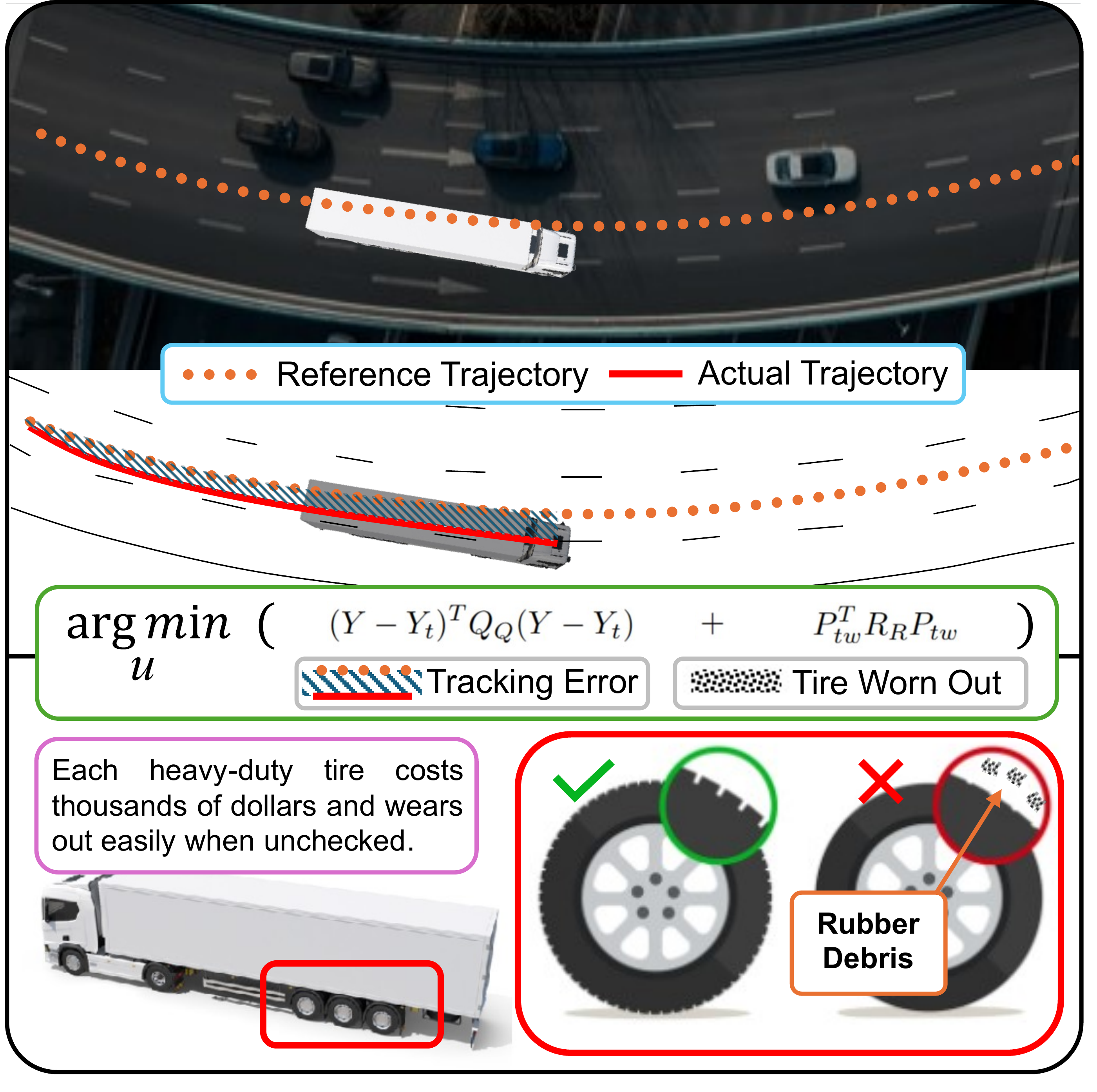}
    \caption{Motivation of the Proposed MS-AGV planning solution.}
    \label{fig:motivation}
\end{figure}

Over the past several years, researchers have dedicated substantial effort to developing advanced control strategies to address the trajectory tracking problem in MS-AGV systems \cite{cao2024learning}. The core technical difficulty lies in managing the steering wheels, as the increased number of state variables \cite{cao2023doublebee} and the dynamic complexity \cite{10965521} of the system make it challenging to predict and control \cite{cao2020online} its behavior effectively.
To simplify the overall system based on kinematics, methods such as Ackerman Steering \cite{chaudhuri2009kinematic,wu2021learn}, Active Front and Rear Steering (AFRS) \cite{ye2016steering}, and Front Wheel Steering (FWS) \cite{xu2022hierarchical} have been proposed. With the assumption that the steering centers are restricted along the first, last, or middle axles, these methods can use fewer states to describe MS-AGV while limiting its maneuverability. In contrast,  {the} rotation center distance-based steering models (D-based) \cite{zhang2016study, zhang2015steering} allow for a more flexible distribution of the steering center, enhancing steering performance. However, transitioning between different steering modes based on road conditions \cite{esfahani2020unsupervised} still introduces transient disturbances.

Furthermore, to improve the understanding of the behavior of MS-AGV during locomotion, various dynamic models with different degrees of freedom have been studied \cite{wang2023multi, gao2014turning, zhang2022dual, hu2016control, wu2017optimizing}. In tire dynamics modeling, some studies rely on small slip assumptions, approximating the relationship between longitudinal/lateral forces and slip ratio/slip angle as linear. This simplification may result in inadequate model representation under high-slip conditions. While a dynamic model presented by Li et al. \cite{li2023control} defines the vehicle's steering center as the intersection point of perpendicular lines to each wheel's longitudinal axis, determined solely by steering angles, it does not account for slip angles. As a result, the actual velocity direction of each wheel deviates from its longitudinal axis. A more accurate representation of the steering center should consider the intersection of perpendicular lines aligned with the actual velocity vectors of the wheels, which are influenced by both steering and slip angles.

Despite the significant costs associated with tire maintenance and replacement, existing studies on MS-AGV trajectory tracking have largely \textbf{overlooked} tire wear reduction. Existing research focuses predominantly on multi-axle tractor trailers \cite{tirewear1,tirewear2,tirewear3,pan2020research,huanjie2014research,du2024hierarchical}, offering insights that are not directly applicable to MS-AGV systems due to differences in structure, operation, and control dynamics.
Most of these studies emphasize tire wear estimation \cite{tirewear1,tirewear2,tirewear3}, optimization models designed specifically for semi-trailers \cite{pan2020research}, and braking strategies to reduce tire wear \cite{huanjie2014research}. In addition, hierarchical coordinated control approaches \cite{du2024hierarchical} attempt to minimize tire wear through isolated optimization processes at different levels. However, such methods often fall short in achieving globally optimal control strategies due to their fragmented and independently managed objectives, highlighting a \textbf{persistent bias} toward localized optimization rather than a comprehensive, system-wide approach.

To enhance the operational durability \cite{li2025ll} of MS-AGV and reduce tire degradation, we propose a novel tire wear-aware trajectory tracking control method. The main contributions of the paper are summarized as
follows:
\begin{itemize}
    \item We introduce the Magic Formula Tire Model at the velocity level, effectively simplifying the representation of tire wear in the context of MS-AGV, avoiding the complexity of direct torque-level modeling in a high-dimensional state and control space.

    \item We formulate a Model Predictive Control (MPC)-based trajectory tracking controller to incorporate tire wear into consideration.  {Our proposed control reduces the state vector size by focusing on steering velocity and wheel speed as control signals, delegating velocity control to a lower-level controller.}

    \item  We validate the proposed approach through comprehensive simulations, demonstrating significant reductions in tire wear while preserving control performance.


\end{itemize}

The rest of the paper is organized as follows: Section \ref{sec:2} provides a detailed dynamics analysis, simplifies the MS-AGV model, and states the problem under study. Section \ref{sec:3} followed by the detailed controller design. In Section \ref{sec:4}, we validate the proposed method by comparing the performance with previous works and provide a comprehensive performance analysis. Section \ref{sec:5} concludes the paper.

\section{MS-AGV Dynamics and Problem Statement}\label{sec:2}

\subsection{Preliminary}
\begin{figure}
    \centering
    \includegraphics[width=1\linewidth]{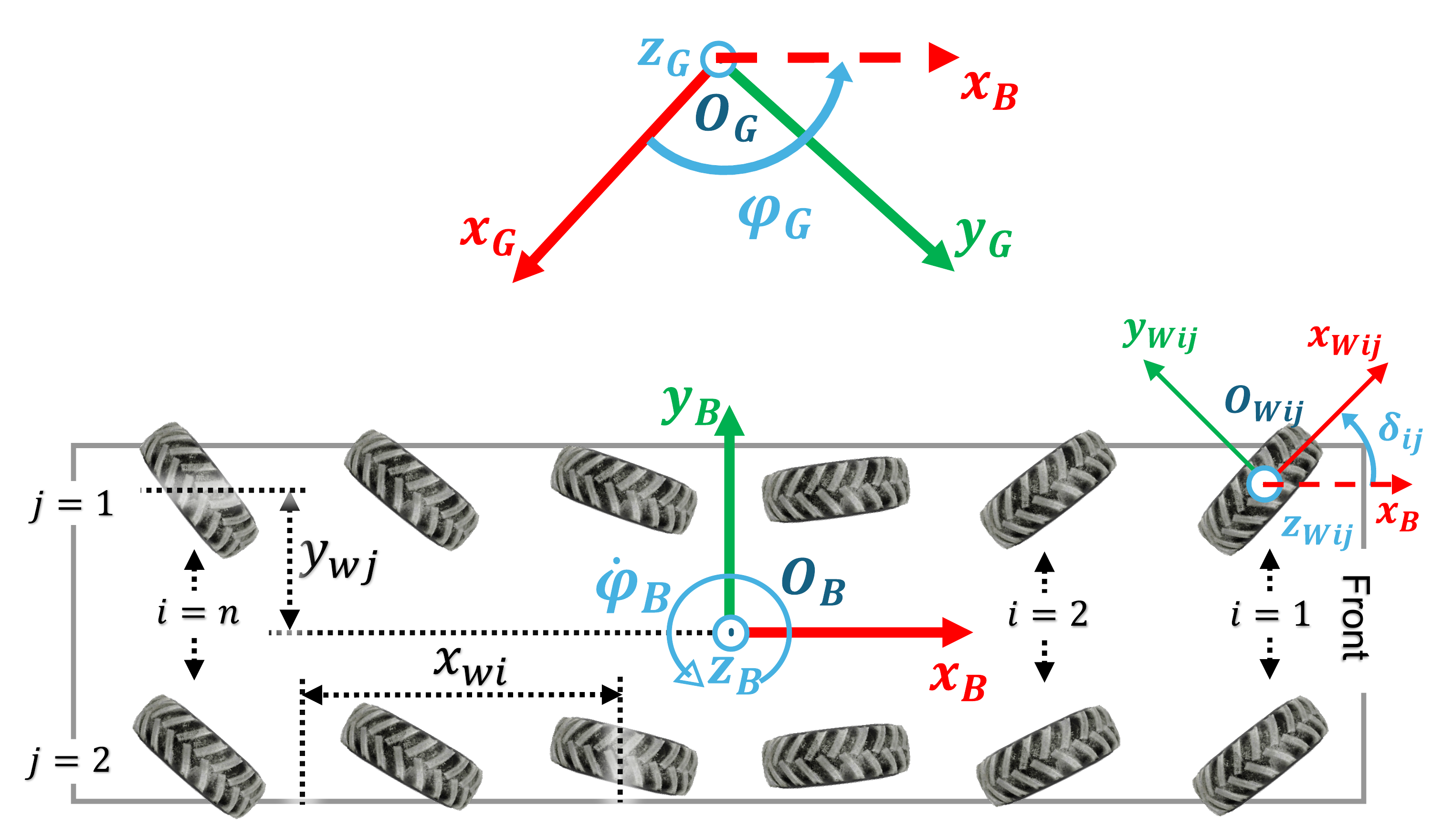}
    \caption{Diagram of MS-AGV Coordinate System}
    \label{fig:Coordinates}
\end{figure}

As shown in Fig~\ref{fig:Coordinates}, MS-AGV's locomotion can be described by using the global coordinate system $\frameglobal=\{\globO, \globXvec, \globYvec, \globZvec\}$, the body coordinate system $\framebody=\{\vehicleO, \vehicleXvec, \vehicleYvec, \vehicleZvec\}$ and the $ij$-th wheel coordinate system $\framewheelij=\{\wheelOij, \wheelXijvec, \wheelYijvec, \wheelZijvec\}$, where $O,x,y,z$ represent the origin and the basis vectors and the subscripts $(\cdot)_\frG, (\cdot)_\frB, (\cdot)_{\frWij}$ indicate the coordinate frame. The term $ij$-th wheel refers to a specific wheel identified by its position indices $i$ and $j$. Here, $i$ indicates the position of the wheel along the $\vehicleXvec$ axis, where $i \in [1, n]$ (with $n$ being the total number of axles of the vehicle), and $j$ specifies the position of the wheel along the $\vehicleYvec$ axis, distinguishing between the left and right wheels, where $j \in \{1, 2\}$.

The body coordinate system $\framebody$ is defined with its origin located at the center of mass of the vehicle. The $\vehicleXvec$ and $\vehicleYvec$ axes are aligned with the vehicle's longitudinal and lateral directions, respectively. Similarly, the wheel coordinate system $\framewheelij$ for the $ij$-th wheel is defined with its origin at the center of mass of the wheel. The $\wheelXijvec$ and $\wheelYijvec$ axes are aligned with the longitudinal and lateral directions of the wheel, respectively. The $\globPhi\in\mathbb{R}$ denotes the angle between the $\globXvec$ axis and the $\vehicleXvec$ axis.

To mathematically describe the orientation relationships between these frames, we define the rotation matrices of the 
 $ij$-th  wheel frame in the vehicle frame and the vehicle frame in the global frame as
 {${}^\frB_\frW \bm{R}_{ij},{}^\frG_\frB \bm{R} \in SO(2)$}. 
These matrices can be expressed as
 {${}^\frB_\frW \bm{R}_{ij} = \bm{R}_z(\wheelyaw)$}
and
 {${}^\frG_\frB \bm{R} = \bm{R}_z(\vehicleYaw)$}
, where $\wheelyaw \in \mathbb{R}$ denotes both the yaw angle of $\framewheelij$ and the steering angle of the $ij$-th wheel, while $\vehicleYaw \in \mathbb{R}$ represents both the yaw angle of $\framebody$ and the vehicle heading angle.  With these notations, the rotation matrix
 {$\bm{R}_z(\cdot)$}
can now be explicitly defined to represent the transformations along the respective $\globZvec$, $\vehicleZvec$ and $\wheelZijvec$ axes, as follows:
\begin{equation}
     {\bm{R}}_z(\psi)=
    \left[\begin{array}
    {ccc}\cos \psi & -\sin \psi\\
    \sin \psi & \cos \psi
    \end{array}\right].
    \label{eq:R}
\end{equation}
Since MS-AGV is commonly employed in indoor environments such as laboratories and factory workshops, as well as outdoor settings like ports, stations, and logistics centers, where the ground is typically well-paved and flat \cite{yuan2024large,Nguyen2025ULOC}.  {In this work, we design a controller for the MS-AGV’s planar motion on flat terrain, assuming that the vehicle maintains zero pitch and roll angles to ensure consistent ground contact.}
Therefore, the following assumption has to be made for a feasible solution.

\begin{Assumption}
\textcolor{black}{The vehicle operates on flat terrain with continuous ground contact}, which means that the vehicle's pitch angle and roll angle are maintained at $0$, constraining the motion to the horizontal plane.
 {This assumption is commonly adopted in similar works, such as \cite{skrickij2024review}.}
\label{assumption:flat}
\end{Assumption}

\subsection{Dynamics of the whole system}
\begin{figure}
    \centering
    \includegraphics[width=1\linewidth]{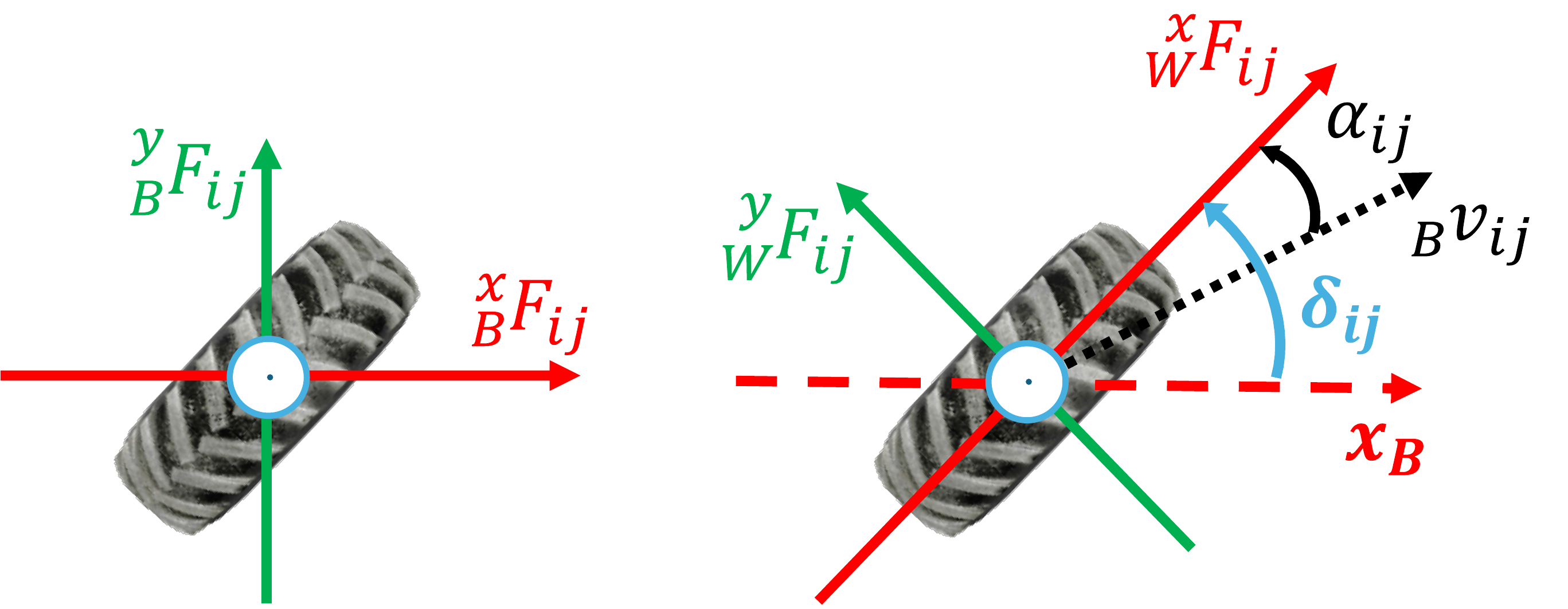}
    \caption{Diagram of MS-AGV Tire Dynamic Model}
    \label{fig:Dynamic}
\end{figure}

Consider the Assumption \ref{assumption:flat}  and the MS-AGV shown in Fig~\ref{fig:Coordinates} and Fig~\ref{fig:Dynamic}, the vehicle has the following relationships as follows:
\begin{equation}
\begin{bmatrix}
    \globAX\\
    \globAY
\end{bmatrix}
=
 {\bm{R}}_z(\vehicleYaw)
\begin{bmatrix}
    \vehicleAX\\
    \vehicleAY
\end{bmatrix},
\end{equation}
\begin{equation}
\globAPhi = \vehiclAPhi,
\end{equation}
where the angular acceleration $ \globAPhi $ is the same in both the global and body coordinate systems.

The longitudinal and lateral forces on tires have the following relationships:
\begin{equation}
\begin{bmatrix}
\FBxij
\vspace{5pt}\\
\FByij
\end{bmatrix}
=
 {\bm{R}}_z(\wheelyaw)
\begin{bmatrix}
\Fwxij
\vspace{5pt}\\
\Fwyij
\end{bmatrix},
\end{equation}
where $\FBxij\in\mathbb{R}$ and $\FByij\in\mathbb{R}$ are the longitudinal and lateral forces in the body coordinate system $\framebody$, respectively. Similarly, the longitudinal and lateral forces in the $ij$-th wheel coordinate system $\framewheelij$ are represented by $\Fwxij$ and $\Fwyij$, respectively. Based on mechanical analysis, the equations of motion for the vehicle are given by:
\begin{equation}
\begin{bmatrix}
m \vehicleAX \\
m \vehicleAY \\
I_{vz} \vehiclAPhi
\end{bmatrix}
=
\sum_{i=1}^{n} 
\sum_{j=1}^{2} 
\begin{bmatrix}
\FBxij \vspace{2pt}\\
\FByij \vspace{2pt}\\
x_{wi} \cdot \FByij - y_{wj} \cdot \FBxij
\end{bmatrix},
\label{eq:dynamicvehicle}
\end{equation}
where the position of $ij$-th wheel in the body coordinate system $\framebody$ is $(x_{wi}, \; y_{wj})$; $m\in\mathbb{R}^+$ is the vehicle mass; $ I_{vz} \in\mathbb{R}^+$ is the vehicle moment of inertia around the $\vehicleZvec$ axis.
\subsection{Problem Statement}

Unlike Ackerman steering vehicles and semi-trailers, the MS-AGV (Multi-axis Steerable Autonomous Mobile Robot) offers superior maneuverability due to its ability to independently control the steering angle and speed of each wheel. However, this capability also presents significant challenges in control. Improper coordination of wheel steering angles and speeds can lead to excessive tire wear, increasing the operational cost of the MS-AGV and introducing potential safety risks. Therefore, optimizing tire wear in the trajectory tracking for MS-AGV is of critical importance.

Given the multi-axis characteristics of MS-AGV, our approach focuses on two key aspects. First, due to the large state space and control vector dimensionality of MS-AGV, directly incorporating tire wear into the torque-level control would result in a highly complex system. To address this, we introduce the Magic Formula Tire Model to represent tire wear at the velocity level, simplifying the modeling process. Second, based on this simplified model, we propose a tire wear-aware trajectory tracking controller using Model Predictive Control (MPC). 
 {In this formulation, steering velocity and wheel speed are used as high-level control inputs instead of low-level steering and wheel torques. These velocity commands are tracked by dedicated low-level controllers, which handle the underlying torque-level execution. This hierarchical decoupling reduces the control input dimension and eliminates the need to model actuator dynamics or torque response in the MPC formulation, thereby significantly simplifying the overall model complexity.}

\section{Tire wear model and Controller Design}\label{sec:3}
\subsection{Dynamics of Tire}
The vehicle velocity is defined as $[\vehicleVX, \vehicleVY, \vehicleVPhi]$, which is projected to the $ij$-th tire location ($x_{wi}, \; y_{wj}$) as
 {$\Vvij$}.
The velocity components in the $\vehicleXvec$ and $\vehicleYvec$ directions are denoted as $\Vvxij$ and $\Vvyij$, are respectively given by:
\begin{equation}
    \Vvxij = \vehicleVX - \vehicleVPhi y_{wj},
    \label{eq:v}
\end{equation}
\begin{equation}
    \Vvyij = \vehicleVY + \vehicleVPhi x_{wi}.
\end{equation}

In the $ij$-th wheel coordinate system $\framewheelij$, the velocity components are transformed by:
\begin{equation}
\begin{bmatrix}
\Vvxij
\vspace{5pt}\\
\Vvyij
\end{bmatrix}
=
 {\bm{R}}(\delta_{ij})
\begin{bmatrix}
\Vwxij
\vspace{5pt}\\
\Vwyij
\end{bmatrix},
\end{equation}

where \(\Vwxij\) and \(\Vwyij\) are the velocity components in the wheel coordinate system $\framewheel$.

The relationship between each wheel's slip angle\cite{pacejka2005tire}, \(\alpha_{ij}\in\mathbb{R}\), and the steering angles $\delta_{ij}$ is given as follows:
\begin{equation}
    \alpha_{ij} = \delta_{ij} - \arctan{\frac{\Vvyij}{\Vvxij}}.
    \label{eq:slip_angle}
\end{equation}
The relationship between $ij$-th wheel's slip ratio \(s_{ij}\in\mathbb{R}\), and the angular velocity $\omegawij\in\mathbb{R}$ is given as follows:
\begin{equation}
    s_{ij} = \frac{\omegawij \cdot r_w - \Vwxij}{\abs{\Vwxij}},
    \label{eq:slip_ratio}
\end{equation}
where \( r_w \) is the wheel radius. Based on the Pacejka Magic Formula tire model\cite{pacejka1992magic}, the tire longitudinal force $\Fwxij$ is related to $s_{ij}$, and the tire lateral force $\Fwyij$ is related to $\alpha_{ij}$. The longitudinal and lateral force can be  {defined} as:
\begin{equation}
\Fwxij = f(s_{ij}),
\label{eq:Fs}
\end{equation}
\begin{equation}
\Fwyij = f(\alpha_{ij}),
\label{eq:Fa}
\end{equation}
where empirical model $f(\cdot)$ is given by:

\[
\begin{split}
   f(\xi) = D \sin \bigg( C \arctan \bigg( B\xi - E \big( B\xi - \arctan(B\xi) \big) \bigg) \bigg).
\end{split}
\]

In the empirical model $f(\cdot)$, the parameter $\xi$ representing the slip ratio $s_{ij}$ or the slip angle $\alpha_{ij}$, is used in conjunction with the following coefficients: the stiffness factor $B\in\mathbb{R}$, the shape factor $C\in\mathbb{R}$, the peak factor $D\in\mathbb{R}$, and the curvature factor $E\in\mathbb{R}$.
\subsection{Tire Wear Model Approximation }
\begin{figure}
    \centering
    \includegraphics[width=1\linewidth]{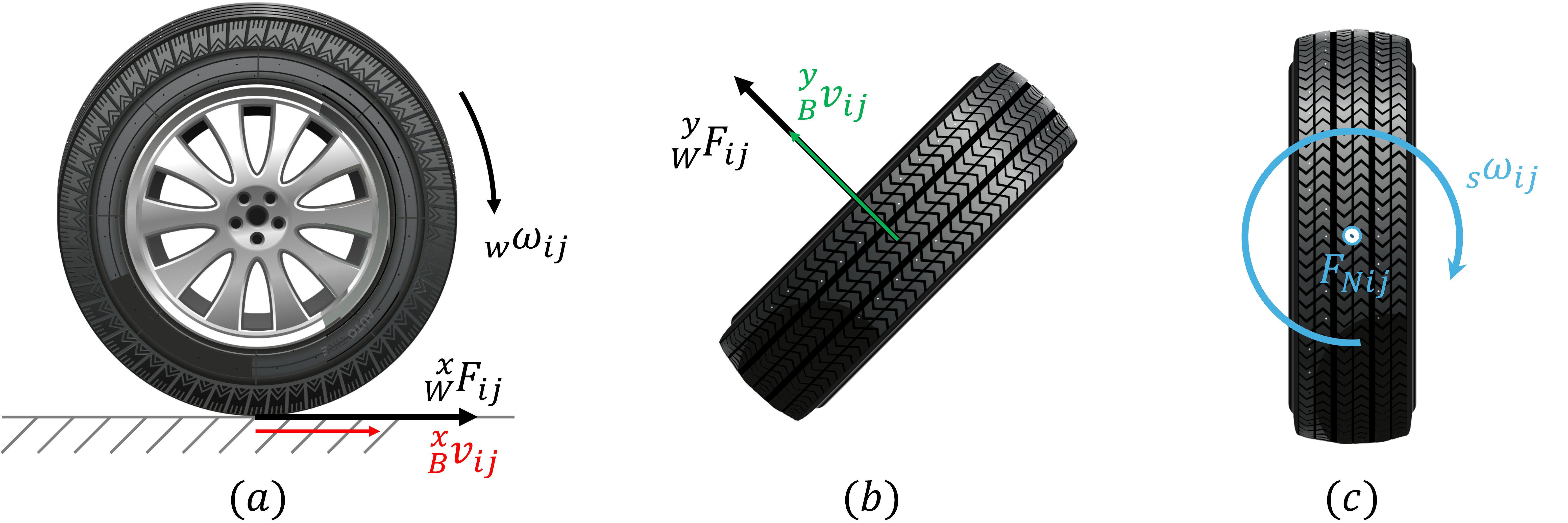}
    \caption{Schematic diagram of tire Wear model}
    \label{fig:tire}
\end{figure}
As shown in Fig~\ref{fig:tire}, Fig~\ref{fig:tire}(a), Fig~\ref{fig:tire}(b) and Fig~\ref{fig:tire}(c) represent the tire wear models for the slip ratio, the slip angle, and the steering, respectively. Following the simplified tire wear model proposed by Da Silva et al.\cite{da2012simplified}, we formulate the power losses associated with these wear mechanisms.

For the slip ratio, its wear power is defined as:
\begin{equation}
    \Psij = \abs{\Fwxij \cdot (\omegawij \cdot r_w - \Vwxij)}.
    \label{eq:PSIJ}
\end{equation}

For the slip angle, its wear power is defined as:
\begin{equation}
    \Palphaij = \abs{\Fwyij \cdot \Vwyij}.
    \label{eq:PAIJ}
\end{equation}

The power loss due to steering is defined as:
\begin{equation}
    \Ptij = \abs{k_t \cdot F_{Nij} \cdot \omegasij},
    \label{eq:PTIJ}
\end{equation}
where \(k_t\) is the loss coefficient for steering, $\omegasij$ is the tire steering angular velocity, and \(F_{Nij}\) is the vertical load on the tire.
 {Since the tire load of heavy vehicles is primarily determined by their own weight and cargo distribution, the effect of dynamic load transfer due to acceleration is relatively small in comparison. Accurately capturing such variations would require modeling suspension dynamics, which adds significant complexity. However, for the purpose of estimating tire wear under realistic operational conditions of a multi-axle heavy-duty vehicle. Therefore, the following assumption has to be made for a feasible solution.}

\begin{Assumption}
\textcolor{black}{The load of each tire $F_{Nij}$ remains constant throughout the motion of the vehicle.}
 {This assumption is commonly adopted in similar works, such as \cite{zhou2023impact}.}
\label{assumption:load}
\end{Assumption}

With Assumption \ref{assumption:load} in mind, we can define the tire wear power \(P_{tw} \in \mathbb{R}^{3\cross1}\) by combining equations (\ref{eq:PSIJ}) with (\ref{eq:PTIJ}). The tire wear power \(P_{tw}\) can be calculated as:
\begin{equation} \label{eq:16}
    P_{tw} = \sum_{i=1}^{n}\sum_{j=1}^{2}
    \begin{bmatrix}
    \Psij\\
    \Palphaij\\
    \Ptij
\end{bmatrix}.
\end{equation}

\subsection{Integrated Dynamic Model}
The vehicle acceleration is defined as $\ddot{\mathcal{X}}=[\globAX, \globAY, \globAPhi] {^\mathrm{T}}$. From Equations (\ref{eq:R}) to (\ref{eq:dynamicvehicle}) and Assumption \ref{assumption:flat}, we can express the vehicle's acceleration in the global coordinate system $\frameglobal$ as follows:

\begin{equation}
\ddot{\mathcal{X}}=
\sum_{i=1}^{n}\sum_{j=1}^{2}
\underbrace{
\begin{bmatrix}
\multicolumn{2}{c}{\frac{ {\bm{R}}(\vehicleYaw)}{m}}  \\[-5pt]
\multicolumn{2}{c}{\rule{1.4cm}{0.4pt}} \\[-2pt]
\frac{-y_{wj}}{I_{vz}} & \frac{x_{wi}}{I_{vz}}
\end{bmatrix}
}_{3 \times 2}
 {\bm{R}}(\wheelyaw)
\begin{bmatrix}
    \Fwxij
    \vspace{5pt}\\
    \Fwyij
\end{bmatrix}.
\label{eq:A}
\end{equation}

Substituting equations (\ref{eq:Fs}) and (\ref{eq:Fa}) into (\ref{eq:A}), the vehicle acceleration can be described as a nonlinear function as:
\begin{equation}
\ddot{\mathcal{X}}
=g(\vehicleYaw, \delta_{ij}, s_{ij}, \alpha_{ij}),
\label{eq:dynamicglobal}
\end{equation}
where 
\begin{equation*}
g(\cdot)=\sum_{i=1}^{n}\sum_{j=1}^{2}
\underbrace{
\begin{bmatrix}
\multicolumn{2}{c}{\frac{ {\bm{R}}(\vehicleYaw)}{m}}  \\[-5pt]
\multicolumn{2}{c}{\rule{1.4cm}{0.4pt}} \\[-2pt]
\frac{-y_{wj}}{I_{vz}} & \frac{x_{wi}}{I_{vz}}
\end{bmatrix}
}_{3 \times 2}
 {\bm{R}}(\wheelyaw)
\begin{bmatrix}
    f(s_{ij})
    \vspace{5pt}\\
    f(\alpha_{ij})
\end{bmatrix}.
\end{equation*}

\subsection{Model Discretization}
\begin{figure*}
    \centering
    \includegraphics[width=1\linewidth]{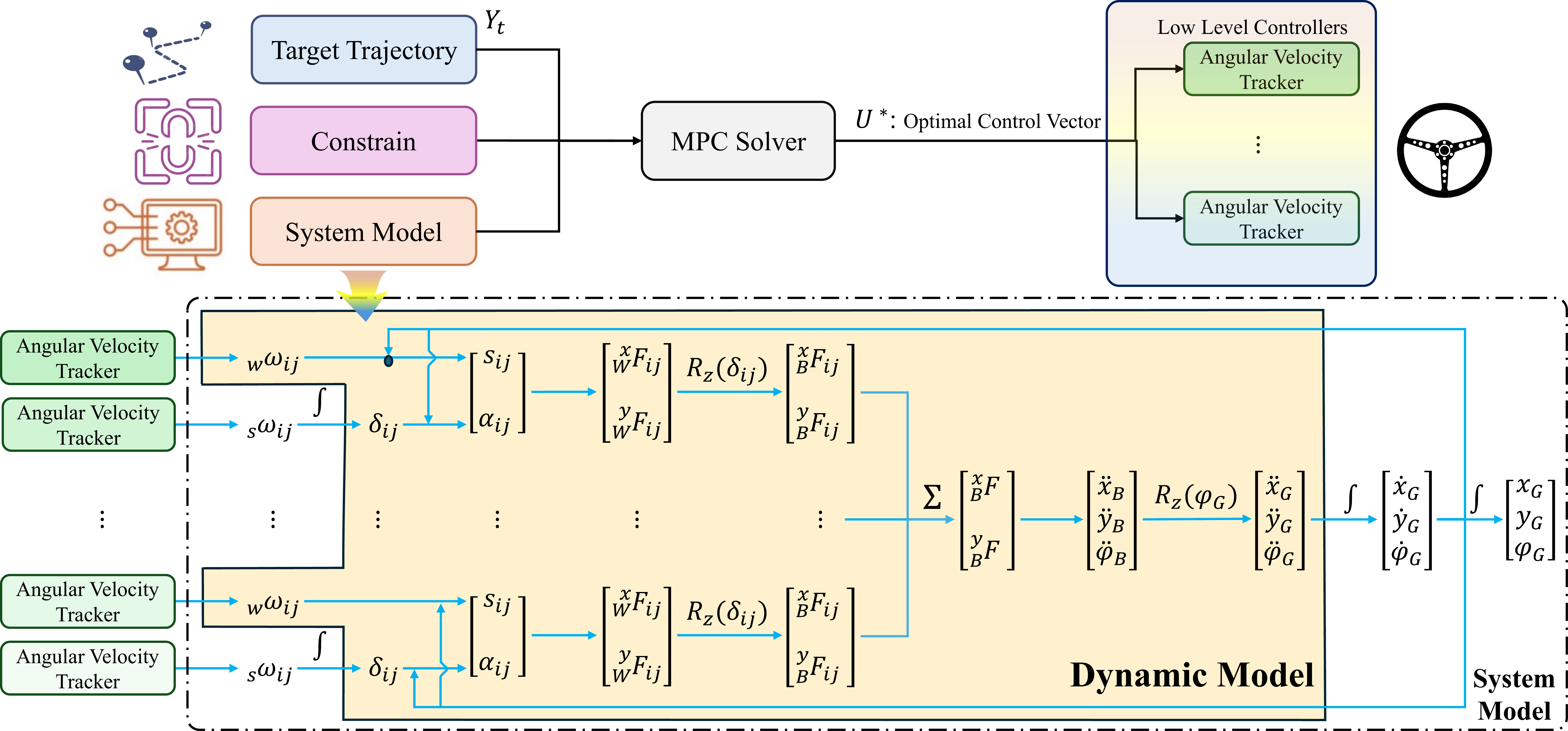}
    \caption{Controller Structure}
    \label{fig:Controller_Structure}
    \vspace{-15pt}
\end{figure*}
As shown in {Fig~\ref{fig:Controller_Structure}}, the system includes an Angular Velocity Tracker, which enables us to design a controller that maps the angular velocities $\omegawij$ and $\omegasij$ to the global position of the vehicle. Specifically, the steering angular velocity $\omegasij$ is used to control the steering angle $\wheelyaw$. Furthermore, vehicle position is defined as $\mathcal{X} =[\globX, \globY, \globPhi] {^\mathrm{T}}$.
To implement this control scheme in a discrete-time system, the forward Euler method is applied for discretization with a time step of \(T\in\mathbb{R}\).  Defined the vehicle velocity as $\dot{\mathcal{X}}=[\globVX,
        \globVY,
        \globVPhi] {^\mathrm{T}}$, the resulting discretized state transition equation is derived as:
\begin{equation}\label{eq:19}
    \wheelyaw(k+1) = \wheelyaw(k) + T \omegasij(k),
\end{equation}
\begin{equation}\label{eq:20}
    \dot{\mathcal{X}}(k+1) = 
      \dot{\mathcal{X}}(k)+ 
    T 
     \ddot{\mathcal{X}}(k),
\end{equation}
\begin{equation}\label{eq:21}
   \mathcal{X}(k+1) = 
     \mathcal{X}(k) + 
    T  
    \dot{\mathcal{X}}(k).
\end{equation}
\begin{Remark}
Equation (\ref{eq:dynamicglobal}) is nonlinear and does not require discretization, as $\ddot{\mathcal{X}}(k)$ can be obtained directly by reformulating it into the following form:
\[
\ddot{\mathcal{X}}(k)
=g(\vehicleYaw(k), \delta_{ij}(k), s_{ij}(k), \alpha_{ij}(k)),
\]
where all parameters can be obtained from the previous state $\mathcal{X}(k)$ and $\dot{\mathcal{X}}(k)$.
\end{Remark}

\subsection{Model Predictive Controller}
The single step control vector $u$ at time $k$ is defined as:
\[
u(k) = [\prescript{}{w}{\omega_{11}}, \dots, \prescript{}{w}{\omega_{n2}},\\ \prescript{}{s}{\omega_{11}}, \dots, \prescript{}{s}{\omega_{n2}}] {^\mathrm{T}}, u \in \mathbb{R}^{4n\cross1}.
\]
\begin{Theorem}
 {
\textbf{Feasibility of Velocity-Level Control Inputs.}
The selection of the control vector $u(k)$, comprising the wheel angular velocities $\omegawij$ and steering angular velocities $\omegasij$ for $ij$-th wheel, is a feasible control strategy for the MS-AGV system.
\\
\textbf{Justification:}
Under the Assumptions~\ref{assumption:flat} and ~\ref{assumption:load}, the vehicle's acceleration, as shown in Equation~\ref{eq:dynamicvehicle}, is determined by the longitudinal forces $\FBxij$ and lateral forces $\FByij$ exerted by each $ij$-th wheel. According to Equations~\ref{eq:slip_angle} through~\ref{eq:Fa}, these forces are functions of the slip ratio $s_{ij}$ and the slip angle $\alpha_{ij}$. These slip parameters, in turn, are influenced by the wheel's angular velocity $\omegawij$ and its steering angle $\delta_{ij}$. Furthermore, the steering angle $\delta_{ij}$ is directly controlled by the steering angular velocity $\omegasij$. Concurrently, the tire wear power, described by Equations~\ref{eq:PSIJ} through~\ref{eq:16}, is also shown to be dependent on these same control inputs: the wheel angular velocities $\omegawij$ and steering angular velocities $\omegasij$. Since both the vehicle's acceleration (and thus its trajectory) and the tire wear power can be influenced by manipulating $\omegawij$ and $\omegasij$, the control vector $u(k)$ provides a feasible means to achieve the desired control objectives.
}
\end{Theorem}
The sequences of the state vector $Y\in\mathbb{R}^{3N_p\cross 1}$ in Equation \eqref{eq:21}, the overall control vector $U\in\mathbb{R}^{4nN_c\cross 1}$, and the tire wear power vector $P\in\mathbb{R}^{3N_p\cross 1}$  of the system in Equation \eqref{eq:16} can be expressed as follows:
\[
Y =[\mathcal{X}(k+1), \; \mathcal{X}(k+2), \; \cdots, \; \mathcal{X}(k+N_p)] {^\mathrm{T}}, 
\]
\[
U =[u(k), \; u(k+1), \; \cdots, \; u(k+N_c-1)] {^\mathrm{T}},
\]
\[
P =[P_{tw}(k+1), \; P_{tw}(k+2), \; \cdots, \; P_{tw}(k+N_p)] {^\mathrm{T}},
\]
where \( N_p\in\mathbb{Z}^+ \) represents the prediction horizon, and \( N_c\in\mathbb{Z}^+ \) represents the control horizon. 
 {It should be noted that the reference trajectory $Y$ provided to the MPC must be pre-generated in the trajectory planning stage with feasible turning radii and curvature, such that they conform to the vehicle's mechanical constraints, including the maximum steering angle and steering rate. This ensures that the yaw angle $\varphi_G$ and related kinematic states remain physically achievable during tracking.
}
To ensure that the vehicle follows the expected trajectory while minimizing the power of tire wear, the cost function is defined as:
\begin{equation}
J = (Y - Y_t) {^\mathrm{T}} W_Y (Y - Y_t) +  {P_{tw}^\mathrm{T}} W_P P_{tw}, J\in\mathbb{R}
\label{eq:J}
\end{equation}
where \(Y_t\) represents the target trajectory sequences; \( W_Y \in\mathbb{R}^{3N_p\cross3N_p}\) and \( W_P\in\mathbb{R}^{3N_p\cross3N_p} \) are weight matrices, expressed as follows:
\begin{equation}
W_Y = E_{N_p} \otimes Q,
\end{equation}
\begin{equation}
W_P = E_{N_p} \otimes L,
\end{equation}
where $\otimes$ represents the Kronecker product, and $E_{N_p}$ is $N_P$-dimensional identity matrix. The matrix \(Q \in\mathbb{R}^{3\cross3}\) and \(L\in\mathbb{R}^{3\cross3}\) are the diagonal matrices, i.e.
\begin{equation}
Q = \mathrm{diag}\{Q_X,Q_Y,Q_\Phi\},
\end{equation}
\begin{equation}
L = \mathrm{diag}\{L_s,L_\alpha,L_t\}.
\end{equation}
In the above equation, the \(Q_X\in\mathbb{R}\), \(Q_Y\in\mathbb{R}\) and \(Q_\Phi\in\mathbb{R}\) are the weight of the error in the $\globX$, $\globY$ and $\globPhi$ direction, respectively;  \(L_s\in\mathbb{R}\), \(L_\alpha\in\mathbb{R}\) and \(L_t\in\mathbb{R}\) are the weight of the tire wear power caused by slip ratio, slip angle and steering, respectively.

To ensure vehicle stability, the ranges of $\omegasij\in\mathbb{R}$, $\omegawij\in\mathbb{R}^+$ and $\wheelyaw\in\mathbb{R}$ must be limited:
\[
    \omegasij \in \left(\prescript{}{s}{\omega_{\min}}, \prescript{}{s}{\omega_{\max}}\right),
\]
\[
    \omegawij \in \left(\prescript{}{w}{\omega_{\min}}, \prescript{}{w}{\omega_{\max}}\right),
\]
\[
    \wheelyaw \in \left(\delta_{\min}, \delta_{\max}\right),
\]
where $\prescript{}{s}{\omega_{\min}}\in\mathbb{R}^-$ and $\prescript{}{s}{\omega_{\max}}\in\mathbb{R}^+$ denote the lower and upper bounds of the steering angular velocity, respectively; $\prescript{}{w}{\omega_{\min}}=0$ and $\prescript{}{w}{\omega_{\max}}\in\mathbb{R}^+$ represent the minimum and maximum wheel angular velocities; and $\delta_{\min}\in\mathbb{R}^-$ and $\delta_{\max}\in\mathbb{R}^+$ specify the mechanical constraints on the steering angle. These bounds are imposed to ensure the physical feasibility of the vehicle's motion and to protect the mechanical components from excessive stress.

In summary, the path tracking problem has now been transformed into an optimization problem:
\begin{equation}
\begin{aligned}
\min_U \quad & J = (Y - Y_t) {^\mathrm{T}} W_Y (Y - Y_t) +  {P_{tw}^\mathrm{T}} W_P P_{tw} \\
\text{s.t.} \quad & \begin{cases}
\ddot{\mathcal{X}}
=g(\vehicleYaw, \delta_{ij}, s_{ij}, \alpha_{ij}),\\
\prescript{}{s}{\omega_{\min}} < \omegasij < \prescript{}{s}{\omega_{\max}}, \\
\prescript{}{w}{\omega_{\min}} < \omegawij < \prescript{}{w}{\omega_{\max}}, \\
\delta_{\min} < \wheelyaw < \delta_{\max}.
\end{cases}
\end{aligned}
\label{eq:opt_constrains}
\end{equation}
 {The cost function combines the trajectory tracking error and the predicted tire wear, weighted by $W_Y$ and $W_P$, respectively. By tuning these weights, the controller can slightly reduce tracking accuracy to significantly reduce tire wear. The optimal solution has a dimension of $U^* \in \mathbb{R}^{4nN_c \times 1}$, where $n$ is the number of axles and $N_c$ is the control horizon.}

\section{Experiment}\label{sec:4}
\subsection{Experiment Setup}
\begin{figure}
    \centering
    \includegraphics[width=1\linewidth]{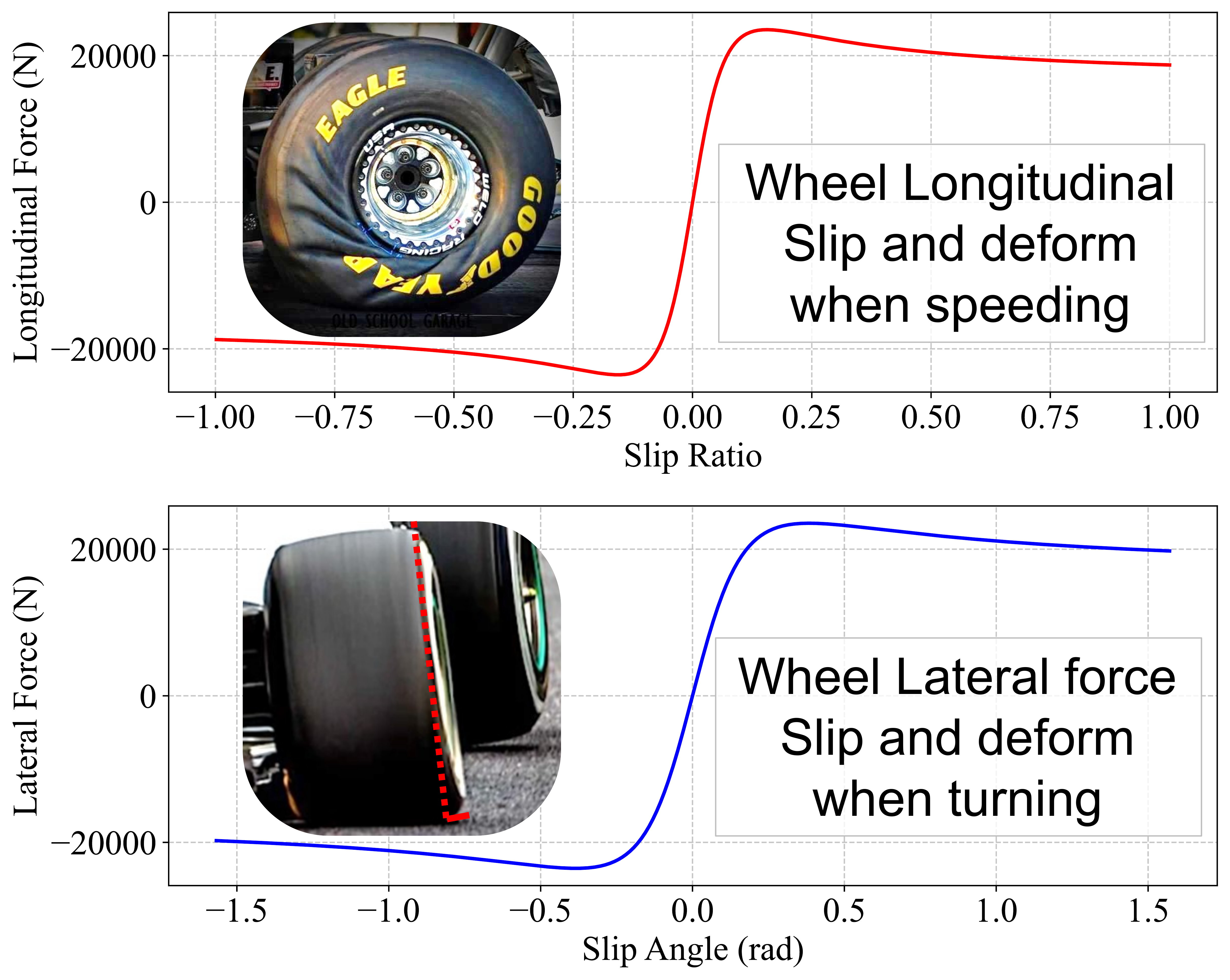}
    \caption{Tire Longitudinal and Lateral Force Characteristics}
    \label{fig:Longitudinal_Lateral}
    \vspace{-15pt}
\end{figure}

This work focuses on trajectory tracking for MS-AGV. The mechanical model is established using the Pacejka Magic Formula tire model. However, due to experimental condition limitations, we constructed a two-axle Swerve-Drive Vehicle and utilized MATLAB simulations to verify the performance. To ensure high simulation accuracy, MATLAB's ODE45 variable-step Runge-Kutta method was employed for numerical integration of the vehicle dynamics. The vehicle has a length of 7 meters and a width of 3 meters, with wheel positions distributed at \(x_{wi} \in \{3, -3\}\) and \(y_{wj} \in \{1, -1\}\). The vehicle moment of inertia around the $\vehicleZvec$ is \(I_{vz} = 80000 \;\text{kg·m}^2\). The tire radius of 0.5 meters and the longitudinal and lateral force characteristics of the tire are shown in {Fig~\ref{fig:Longitudinal_Lateral}}. The trajectory tracking control is implemented using a Model Predictive Control (MPC) framework with simulated annealing optimization. 

\subsection{ {Baselines Selections}}

 {
We evaluate five representative vehicle modeling approaches within a unified trajectory tracking framework, with a focus on assessing their effectiveness in balancing tracking accuracy and tire wear performance. The selected models are chosen based on their relevance to trajectory control tasks, availability of implementation, and compatibility with multi-axle swerve-drive AGVs. While many related works exist, several are excluded from direct comparison due to fundamentally different assumptions or system objectives. For example, the Prescribed Performance Fault-Tolerant Control method~\cite{zhang2024fault} is designed for scenarios involving actuator failures, which fall outside the scope of our study that assumes nominal operation. Similarly, the neural control strategy proposed for underactuated marine vehicles~\cite{zhang2022neural} targets platforms with strong model uncertainty and partial actuation, in contrast to our fully actuated ground vehicle setting. Additionally, the trajectory planning approach for minimizing tire wear in articulated vehicles~\cite{papaioannou2024reducing} addresses a different vehicle type and focuses on offline planning, whereas our work centers on real-time trajectory tracking for swerve-drive AGVs. Ultimately, we selected the following models for quantitative comparison.}

 {
\noindent \textbf{Kinematic Model (Kinematic):} A kinematic modeling approach for MS-AGV~\cite{zhang2016study} that considers only geometric relationships without accounting for tire slip or wear effects. This model represents a widely used baseline in path tracking tasks due to its simplicity and low computational load, despite lacking dynamic fidelity.}

 {
\noindent \textbf{Dynamic Model by Du et al. (DH):} A vehicle dynamic model proposed by Du et al.~\cite{du2024hierarchical} that incorporates the Dugoff tire model for tire-road interaction calculations. This model serves as a representative of dynamic modeling with simplified tire dynamics, enabling fair comparison with our Pacejka-based formulation.
}

 {
\noindent \textbf{Efficient Dynamic Model by Tang et al. (TCT):} A vehicle dynamic model developed by Tang et al.~\cite{tang2024all} that models only the lateral ($\vehicleYvec$) and yaw ($\vehicleYaw$) dynamics for computational efficiency, while the longitudinal motion ($\vehicleXvec$) is controlled with a constant velocity assumption. It reflects a common trade-off strategy in the literature where only partial dynamics are considered to reduce computational cost.
}

 {
\noindent \textbf{Our Proposed Model Without Tire Wear Optimization (NTWO):} Our proposed vehicle dynamic model using the Pacejka Magic Formula tire model, with the tire wear weight matrix $W_P$ set to zero to focus purely on tracking performance. This variant allows us to isolate the effect of tire wear optimization by keeping all other modeling components consistent.
}

 {
\noindent \textbf{Our Proposed Full Model With Tire Wear Optimization (TWO):} Our proposed vehicle model incorporates both vehicle dynamics and tire wear optimization through appropriate weighting in $W_P$ to achieve a balance between tracking accuracy and tire longevity. This is the full version of our approach, designed to reflect realistic AGV operational conditions with an explicit focus on tire health and maintenance cost.
}

\subsection{Evaluation Metric}
The evaluation metrics for our work are listed as follows:
\begin{itemize}
    \item \textbf{Tire Wear Work Due to Slip Ratio \textbf{$W_{s} \in \mathbb{R}$}:} The total tire wear work attributed to the slip ratio is defined as:
    \begin{equation}
        W_{s} = \sum_{n=1}^{N}P_{sn} \cdot T, 
    \label{eq:WS}
    \end{equation}
    where $P_{sn} \in \mathbb{R}$ is the $n$-th step tire wear power due to slip ratio, $T$ is the control period, and N is the total number of control steps.
    \item \textbf{Tire Wear Work Due to Slip Angle \textbf{$W_{\alpha} \in \mathbb{R}$}:} The total tire wear work attributed to the slip angle is defined as:
    \begin{equation}
        W_{\alpha} = \sum_{n=1}^{N_p}P_{\alpha n} \cdot T,
    \label{eq:WA}
    \end{equation}
    where $P_{\alpha n} \in \mathbb{R}$ is the $n$-th step tire wear power due to slip angle.
    \item \textbf{Tire Wear Work Due to Steering \textbf{$W_{t}$}:} The total tire wear work attributed to the steering is defined as:
    \begin{equation}
        W_{t} = \sum_{n=1}^{N_p}P_{tn} \cdot T,
    \label{eq:WT}
    \end{equation}
    where $P_{twn} \in \mathbb{R}$ is the n-th step tire wear power due to steering.
    
    \item \textbf{Total Tire Wear Work \textbf{$W_{tw} \in \mathbb{R}$}:} The sum of work done by wear caused by slip angle, slip ratio, and steering, defined as follows:
    \begin{equation}
        W_{tw} = W_{s} + W_{\alpha} + W_{t},
    \label{eq:WTW}
    \end{equation}
    \item \textbf{RMSE in each Direction $e_x \in \mathbb{R}$, $e_y \in \mathbb{R}$, $e_{\varphi} \in \mathbb{R}$:} Root mean square errors in X, Y, and heading directions, respectively.
    \item \textbf{Average RMSE $\Bar{e} \in \mathbb{R}$:} The arithmetic mean of $e_x$, $e_y$ and $e_{\varphi}$, defined as follows:
    \begin{equation}
        \Bar{e} = \frac{e_x + e_y + e_{\varphi}}{3},
    \label{eq:AvgE}
    \end{equation}
    \item \textbf{Performance Balance Index $\Omega \in \mathbb{R}$:} A comprehensive evaluation metric that characterizes the trade-off between tire wear work and tracking accuracy, defined as follows:
    \begin{equation}
        \Omega =
        \begin{cases} 
        (W_{tw})^{0.5} \cdot (\Bar{e})^{0.1}, & \text{if } \Bar{e} < 50, \\
        (W_{tw})^{0.5} \cdot (\Bar{e})^{0.5}, & \text{if } \Bar{e} \geq 50.
        \end{cases},
        \label{eq:AvgF}
    \end{equation}

\end{itemize}
\subsection{Results and Discussion}

\subsubsection{Case 1}
\begin{table*}[t]
\centering
\setlength{\tabcolsep}{3pt} 
\renewcommand{\arraystretch}{1.5} 
\caption{Comparison of Metrics in Case 1. Best results are in \textbf{bold}, second best are \underline{underlined}. }
\begin{tabular}{c|l|c|cccc|cccc}
\hline
\hline
& \textbf{Method}
& \textbf{$\Omega$}
& \textbf{$W_{tw}$ ($J$)}
& \textbf{$W_{\alpha}$ ($J$)}
& \textbf{$W_{s}$ ($J$)}
& \textbf{$W_{t}$ ($J$)}
& \textbf{$\Bar{e}$}
& \textbf{$e_x$ ($cm$)}
& \textbf{$e_y$ ($cm$)}
& \textbf{$e_{\varphi}$ (°)} \\ \hline 

\multirow{5}{*}{\rotatebox{90}{35 km/h}}
& Kinematic\cite{zhang2016study}
& \cellcolor[HTML]{D5F5E3} \underline{3.73 × 10\textsuperscript{5}}  
& \cellcolor[HTML]{A9DFBF} \textbf{3.16 × 10\textsuperscript{8}}  
& \cellcolor[HTML]{A9DFBF} \textbf{8.98 × 10\textsuperscript{7}}  
& \cellcolor[HTML]{A9DFBF} \textbf{1.45 × 10\textsuperscript{8}}  
& 8.04 × 10\textsuperscript{7}  
& 440.77  
& 906.76  
& 354.23  
& 61.31  \\

& DH\cite{du2024hierarchical}
& 4.81 × 10\textsuperscript{4}  
& 1.07 × 10\textsuperscript{9}  
& 3.01 × 10\textsuperscript{8}  
& 7.56 × 10\textsuperscript{8}  
& 7.97 × 10\textsuperscript{6}  
& 48.74  
& 55.50  
& 50.86  
& 39.85  \\

& TCT\cite{tang2024all}
& 6.82 × 10\textsuperscript{4}  
& 2.20 × 10\textsuperscript{9}  
& 2.68 × 10\textsuperscript{8}  
& 1.93 × 10\textsuperscript{9}  
& 7.79 × 10\textsuperscript{6}  
& 42.40  
& 43.26  
& 55.40  
& 28.54  \\

& NTWO (Our)
& 3.79 × 10\textsuperscript{4}  
& 9.12 × 10\textsuperscript{8}  
& 2.12 × 10\textsuperscript{8}  
& 6.92 × 10\textsuperscript{8}  
& \cellcolor[HTML]{A9DFBF} \textbf{7.51 × 10\textsuperscript{6}}  
& \cellcolor[HTML]{A9DFBF} \textbf{9.60}  
& \cellcolor[HTML]{D5F5E3} \underline{12.10}  
& \cellcolor[HTML]{A9DFBF} \textbf{8.68}  
& \cellcolor[HTML]{A9DFBF} \textbf{8.03}  \\

& TWO (Our)
& \cellcolor[HTML]{A9DFBF} \textbf{3.63 × 10\textsuperscript{4}}  
& \cellcolor[HTML]{D5F5E3} \underline{7.37 × 10\textsuperscript{8}}  
& \cellcolor[HTML]{D5F5E3} \underline{2.27 × 10\textsuperscript{8}}  
& \cellcolor[HTML]{D5F5E3} \underline{5.03 × 10\textsuperscript{8}}  
& \cellcolor[HTML]{D5F5E3} \underline{7.59 × 10\textsuperscript{6}}  
& \cellcolor[HTML]{D5F5E3} \underline{18.44}  
& \cellcolor[HTML]{A9DFBF} \textbf{9.58}  
& \cellcolor[HTML]{D5F5E3} \underline{27.51}  
& \cellcolor[HTML]{D5F5E3} \underline{18.23}  \\ \hline

\multirow{5}{*}{\rotatebox{90}{25 km/h}}
& Kinematic\cite{zhang2016study}
& 3.31 × 10\textsuperscript{5}  
& \cellcolor[HTML]{A9DFBF} \textbf{5.82 × 10\textsuperscript{8}}  
& \cellcolor[HTML]{A9DFBF} \textbf{1.38 × 10\textsuperscript{8}}  
& \cellcolor[HTML]{A9DFBF} \textbf{2.36 × 10\textsuperscript{8}}  
& 2.08 × 10\textsuperscript{8}  
& 290.47  
& 568.62  
& 245.02  
& 57.79  \\

& DH\cite{du2024hierarchical}
& 5.38 × 10\textsuperscript{4}  
& 1.96 × 10\textsuperscript{9}  
& 4.35 × 10\textsuperscript{8}  
& 1.51 × 10\textsuperscript{9}  
& 1.57 × 10\textsuperscript{7}  
& 7.20  
& 10.36  
& 7.75  
& 3.49  \\

& TCT\cite{tang2024all}
& 4.44 × 10\textsuperscript{4}  
& 1.18 × 10\textsuperscript{9}  
& 4.61 × 10\textsuperscript{8}  
& \cellcolor[HTML]{D5F5E3} \underline{7.02 × 10\textsuperscript{8}}  
& 1.54 × 10\textsuperscript{7}  
& 13.89  
& 26.31  
& 14.01  
& \cellcolor[HTML]{D5F5E3} \underline{1.36}  \\

& NTWO (Our)
& \cellcolor[HTML]{D5F5E3} \underline{3.98 × 10\textsuperscript{4}}  
& 1.53 × 10\textsuperscript{9}  
& 2.00 × 10\textsuperscript{8}  
& 1.32 × 10\textsuperscript{9}  
& \cellcolor[HTML]{D5F5E3} \underline{1.53 × 10\textsuperscript{7}}  
& \cellcolor[HTML]{A9DFBF} \textbf{1.22}  
& \cellcolor[HTML]{A9DFBF} \textbf{1.66}  
& \cellcolor[HTML]{A9DFBF} \textbf{1.26}  
& \cellcolor[HTML]{A9DFBF} \textbf{0.74}  \\

& TWO (Our)
& \cellcolor[HTML]{A9DFBF} \textbf{3.46 × 10\textsuperscript{4}}  
& \cellcolor[HTML]{D5F5E3} \underline{1.01 × 10\textsuperscript{9}}  
& \cellcolor[HTML]{D5F5E3} \underline{1.76 × 10\textsuperscript{8}}  
& 8.19 × 10\textsuperscript{8}  
& \cellcolor[HTML]{A9DFBF} \textbf{1.49 × 10\textsuperscript{7}}  
& \cellcolor[HTML]{D5F5E3} \underline{2.52}  
& \cellcolor[HTML]{D5F5E3} \underline{3.04}  
& \cellcolor[HTML]{D5F5E3} \underline{2.27}  
& 2.25  \\ \hline

\multirow{5}{*}{\rotatebox{90}{15 km/h}}
& Kinematic\cite{zhang2016study}
& 4.03 × 10\textsuperscript{5}  
& \cellcolor[HTML]{A9DFBF} \textbf{1.69 × 10\textsuperscript{9}}  
& \cellcolor[HTML]{A9DFBF} \textbf{3.04 × 10\textsuperscript{8}}  
& \cellcolor[HTML]{A9DFBF} \textbf{6.32 × 10\textsuperscript{8}}  
& 7.56 × 10\textsuperscript{8}  
& 95.74  
& 156.65  
& 90.96  
& 39.62  \\

& DH\cite{du2024hierarchical}
& 9.06 × 10\textsuperscript{4}  
& 5.47 × 10\textsuperscript{9}  
& 8.21 × 10\textsuperscript{8}  
& 4.60 × 10\textsuperscript{9}  
& 4.25 × 10\textsuperscript{7}  
& 7.68  
& 10.26  
& 9.21  
& 3.55  \\

& TCT\cite{tang2024all}
& 1.24 × 10\textsuperscript{5}  
& 9.04 × 10\textsuperscript{9}  
& 5.48 × 10\textsuperscript{8}  
& 8.45 × 10\textsuperscript{9}  
& \cellcolor[HTML]{D5F5E3} \underline{4.22 × 10\textsuperscript{7}}  
& 14.05  
& 26.57  
& 14.39  
& \cellcolor[HTML]{D5F5E3} \underline{1.18}  \\

& NTWO (Our)
& \cellcolor[HTML]{D5F5E3} \underline{7.04 × 10\textsuperscript{4}}  
& 5.45 × 10\textsuperscript{9}  
& 8.46 × 10\textsuperscript{8}  
& 4.56 × 10\textsuperscript{9}  
& 4.28 × 10\textsuperscript{7}  
& \cellcolor[HTML]{A9DFBF} \textbf{0.62}  
& \cellcolor[HTML]{A9DFBF} \textbf{0.62}  
& \cellcolor[HTML]{A9DFBF} \textbf{0.68}  
& \cellcolor[HTML]{A9DFBF} \textbf{0.57}  \\

& TWO (Our)
& \cellcolor[HTML]{A9DFBF} \textbf{5.33 × 10\textsuperscript{4}}  
& \cellcolor[HTML]{D5F5E3} \underline{2.42 × 10\textsuperscript{9}}  
& \cellcolor[HTML]{D5F5E3} \underline{3.64 × 10\textsuperscript{8}}  
& \cellcolor[HTML]{D5F5E3} \underline{2.02 × 10\textsuperscript{9}}  
& \cellcolor[HTML]{A9DFBF} \textbf{4.10 × 10\textsuperscript{7}}  
& \cellcolor[HTML]{D5F5E3} \underline{2.24}  
& \cellcolor[HTML]{D5F5E3} \underline{2.63}  
& \cellcolor[HTML]{D5F5E3} \underline{1.97}  
& 2.11  \\ \hline

\multirow{5}{*}{\rotatebox{90}{5 km/h}}
& Kinematic\cite{zhang2016study}
& 9.83 × 10\textsuperscript{5}
& \cellcolor[HTML]{A9DFBF} \textbf{1.76 × 10\textsuperscript{10}}
& \cellcolor[HTML]{A9DFBF} \textbf{1.66 × 10\textsuperscript{9}}  
& \cellcolor[HTML]{A9DFBF} \textbf{3.76 × 10\textsuperscript{9}}  
& 1.22 × 10\textsuperscript{10}  
& 54.92  
& 90.11  
& 51.97  
& 22.70  \\

& DH\cite{du2024hierarchical}
& 3.05 × 10\textsuperscript{5}  
& 5.12 × 10\textsuperscript{10}  
& 2.58 × 10\textsuperscript{9}  
& 4.83 × 10\textsuperscript{10}  
& 3.82 × 10\textsuperscript{8}  
& 20.03  
& 31.57  
& 17.23  
& 11.30  \\

& TCT\cite{tang2024all}
& 9.98 × 10\textsuperscript{6}  
& 8.30 × 10\textsuperscript{11}  
& \cellcolor[HTML]{D5F5E3} \underline{1.91 × 10\textsuperscript{9}}  
& 8.28 × 10\textsuperscript{11}  
& 3.78 × 10\textsuperscript{8}  
& 120.01  
& 224.44  
& 116.25  
& 19.33  \\

& NTWO (Our)
& \cellcolor[HTML]{D5F5E3} \underline{2.43 × 10\textsuperscript{5}}
& 4.86 × 10\textsuperscript{10}  
& 2.87 × 10\textsuperscript{9}  
& 4.56 × 10\textsuperscript{10}  
& \cellcolor[HTML]{A9DFBF} \textbf{1.87 × 10\textsuperscript{8}}  
& \cellcolor[HTML]{A9DFBF} \textbf{2.59}  
& \cellcolor[HTML]{D5F5E3} \underline{3.41}  
& \cellcolor[HTML]{A9DFBF} \textbf{2.90}  
& \cellcolor[HTML]{A9DFBF} \textbf{1.46}  \\

& TWO (Our)
& \cellcolor[HTML]{A9DFBF} \textbf{2.00 × 10\textsuperscript{5}}  
& \cellcolor[HTML]{D5F5E3} \underline{3.17 × 10\textsuperscript{10}}  
& 2.66 × 10\textsuperscript{9}  
& \cellcolor[HTML]{D5F5E3} \underline{2.88 × 10\textsuperscript{10}}  
& \cellcolor[HTML]{D5F5E3} \underline{1.88 × 10\textsuperscript{8}}  
& \cellcolor[HTML]{D5F5E3} \underline{3.25}  
& \cellcolor[HTML]{A9DFBF} \textbf{2.81}  
& \cellcolor[HTML]{D5F5E3} \underline{3.00}  
& \cellcolor[HTML]{D5F5E3} \underline{3.94}  \\ \hline

\end{tabular}
\label{tab:comparison1}
\vspace{3pt}
\end{table*}

\begin{figure}
    \centering
    \includegraphics[width=1\linewidth]{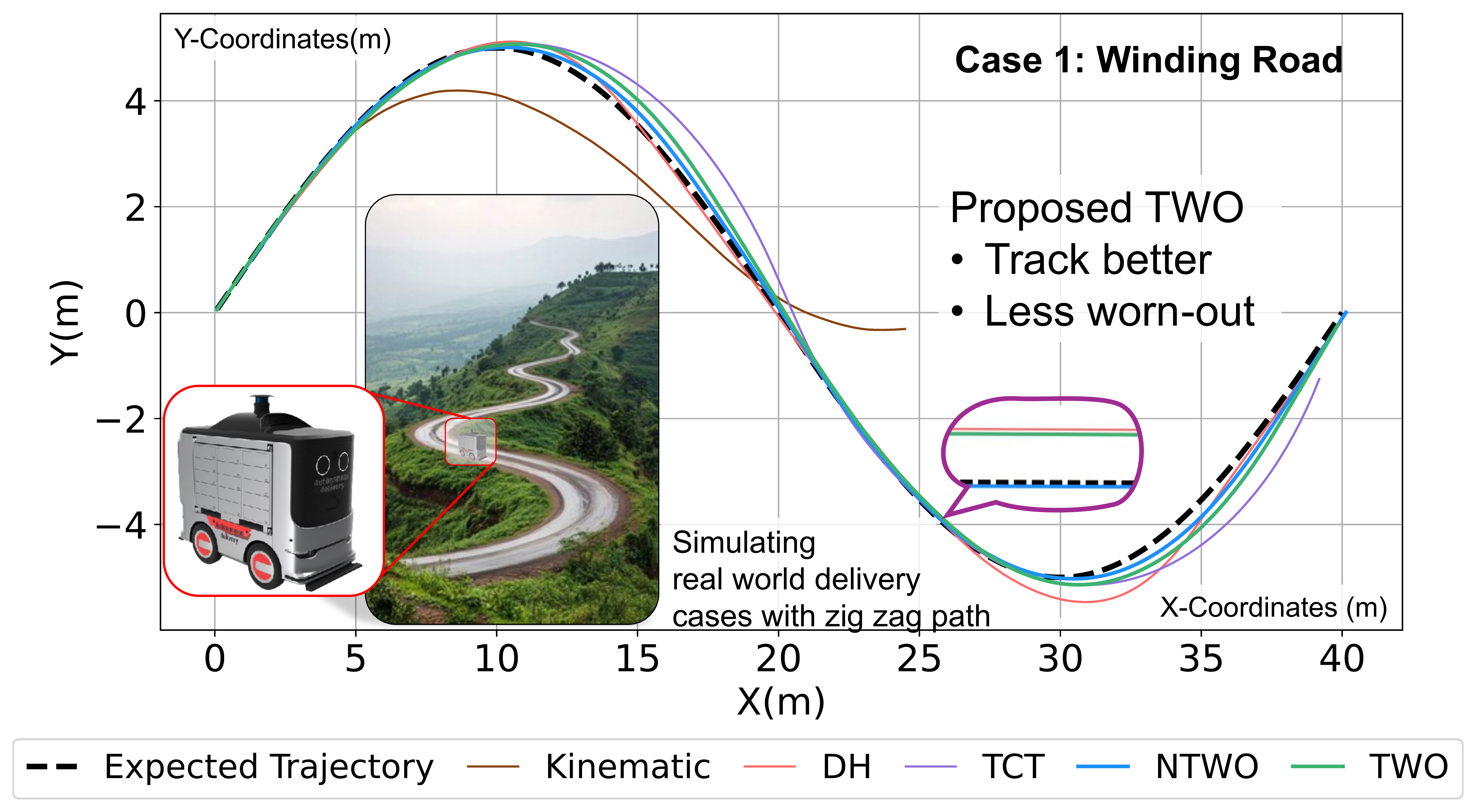}
    \caption{Trajectory Comparison in Case 1}
    \label{fig:TrajectoryComparison1_1}
    \vspace{-15pt}
\end{figure}
\begin{figure}
    \centering
    \includegraphics[width=1\linewidth]{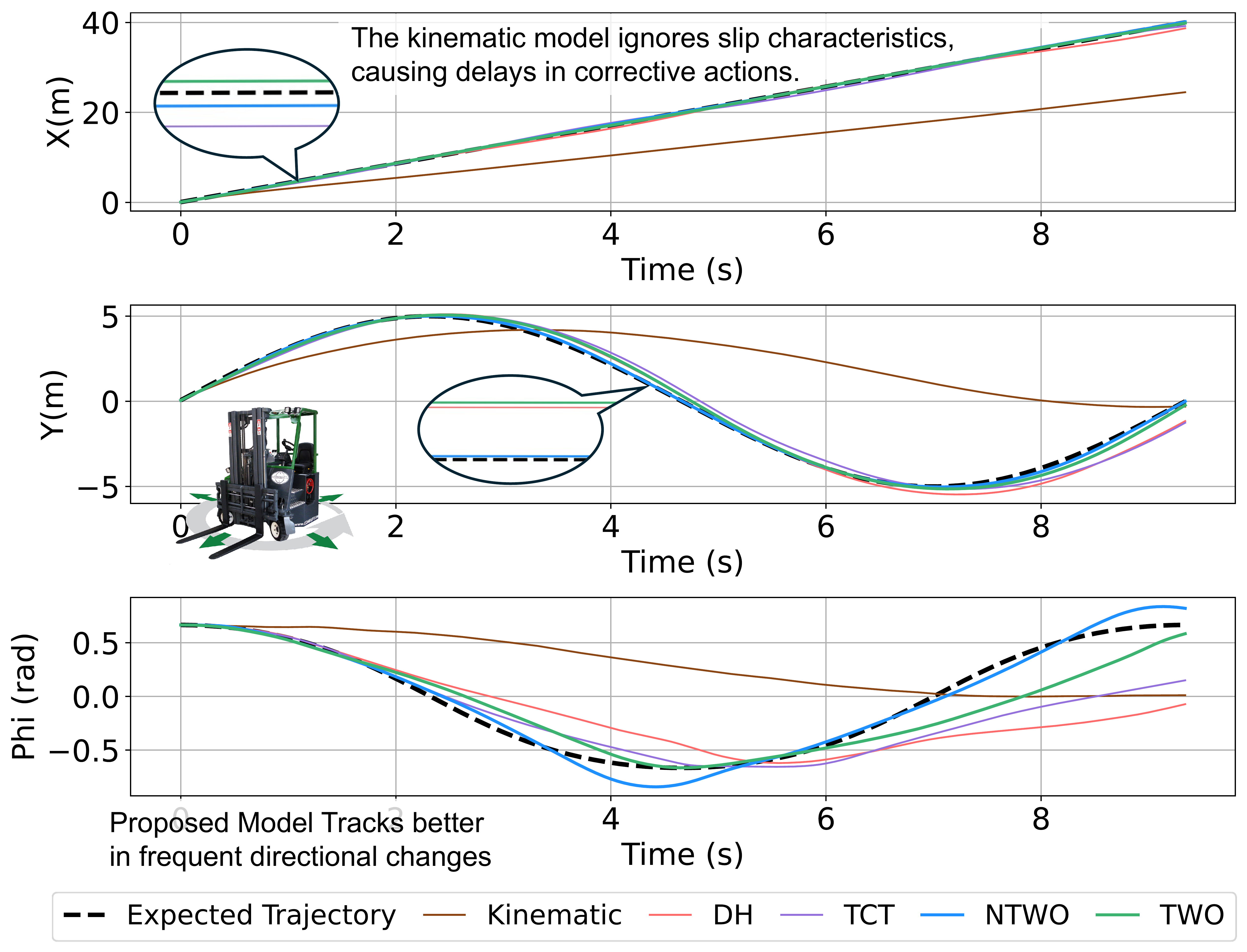}
    \caption{Trajectory Comparison with Time in Case 1}
    \label{fig:TrajectoryComparison1_2}
\end{figure}
\begin{figure}
    \centering
    \includegraphics[width=1\linewidth]{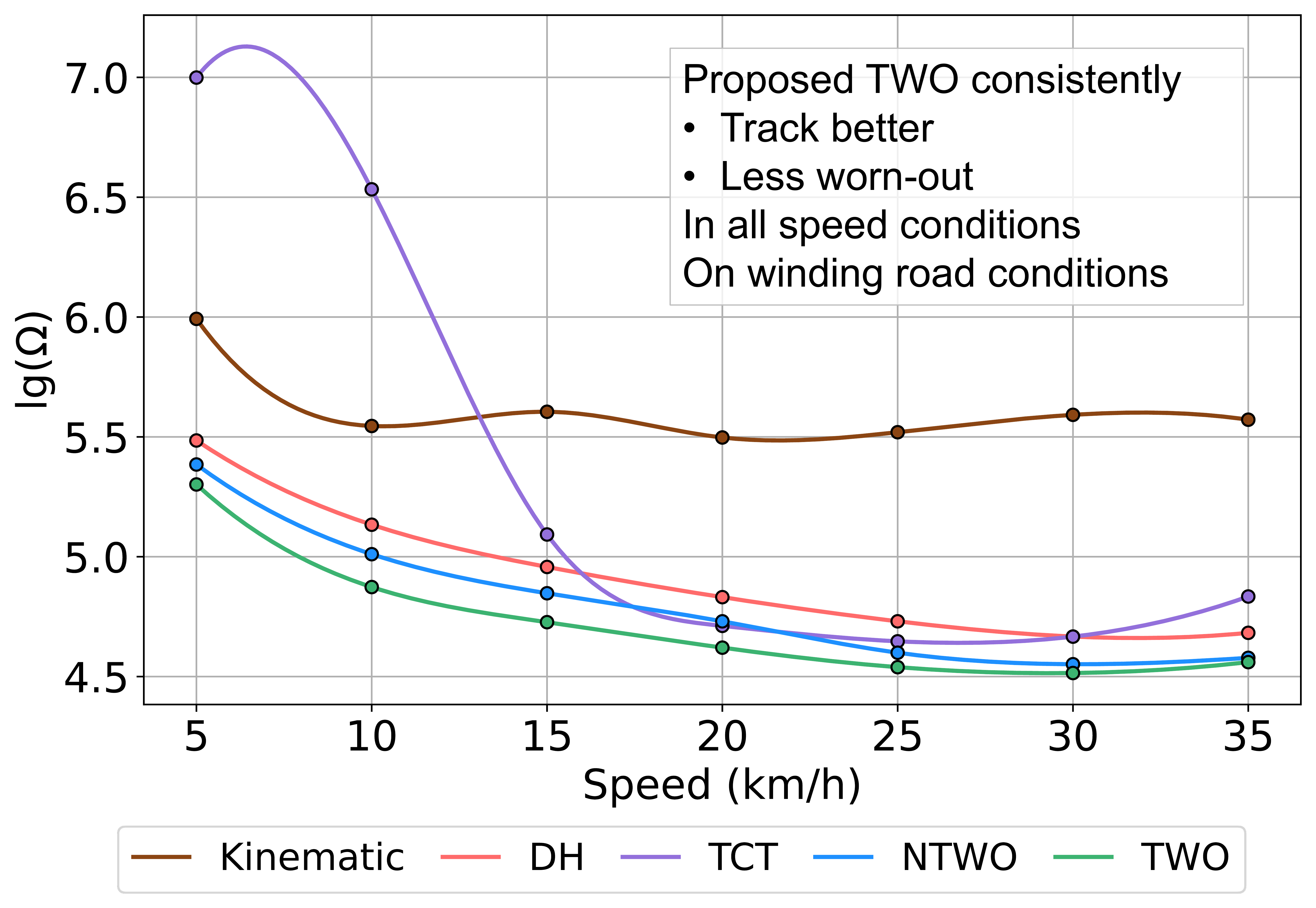}
    \caption{Performance Balance Comparison in Case 1}
    \label{fig:TrajectoryComparison1_3}
    \vspace{-15pt}
\end{figure}

In this case study, a curved path emulating winding road conditions is implemented to analyze trajectory tracking performance, with a gross vehicle weight of 12 t, operating speeds of 5-35 km/h, and a control period of T = 0.01 s. The comparison of trajectory tracking performance is illustrated in {Fig~\ref{fig:TrajectoryComparison1_1}}, {Fig~\ref{fig:TrajectoryComparison1_2}} and {Fig~\ref{fig:TrajectoryComparison1_3}}, while the accumulated tire wear measurements are summarized in {Table \ref{tab:comparison1}}. The tire wear works were evaluated numerically using Equations (\ref{eq:WS}) to (\ref{eq:WTW}).
The results demonstrate that the NTWO achieves the minimum trajectory tracking error. When tire wear optimization (TWO) is incorporated, tire wear is reduced across all speeds, albeit with a slight increase in tracking error. For instance, at 35 km/h, the total tire wear work \textbf{$W_{tw}$} decreases from 9.12 × 10\textsuperscript{8} J to 7.37 × 10\textsuperscript{8} J, representing a 19.19\% reduction. While the Average RMSE $\Bar{e}$ increases from 9.60 to 18.44, this value remains well within acceptable limits. Additionally, the Performance Balance Index $\Omega$ shows a corresponding decrease. This comparison validates that the TWO effectively reduces tire wear work while maintaining acceptable tracking accuracy with moderate performance trade-offs.

In contrast, the Kinematic Method exhibits substantial tracking errors, especially during high-speed curved trajectories, due to its inability to account for tire dynamics, particularly slip. This highlights the advantage of the TWO, which incorporates tire slip characteristics, over the Kinematic Method in high-speed cornering scenarios. By incorporating tire slip dynamics, the Kinematic Method transitions into a Dynamic Model, resembling our Proposed Method. 

\begin{figure}
    \centering
    \includegraphics[width=1\linewidth]{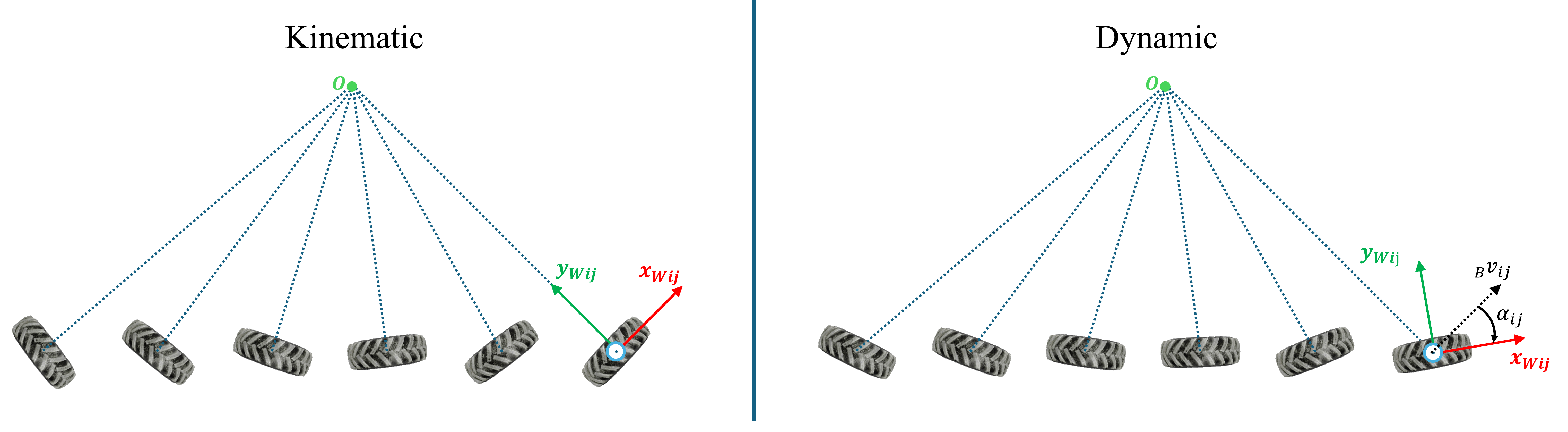}
    \caption{Comparison of Steering Center}
    \label{fig:SteeringCenter}
    \vspace{-15pt}
\end{figure}

However, tire wear in the Kinematic Method is the lowest compared to all other Dynamic Methods. This is because, in the Kinematic Method, the vehicle's steering center is defined as the intersection point of the lateral directions $\wheelYijvec$ of all wheels as shown in Fig~\ref{fig:SteeringCenter}. This ensures that during motion, the individual steering centers of each wheel align with the overall steering center of the vehicle, thereby minimizing slip angle and slip ratio to reducing $W_{tw}$. In contrast, the steering center in all other Dynamic Methods is not defined as the intersection of the lateral directions $\wheelYijvec$, which results in the wheel directions $\wheelXijvec$ deviating from the vehicle's forward direction
 {$\Vvij$},
thereby creating slip angles. While this increases tire wear, the introduction of slip angles allows for a more accurate representation of the vehicle's dynamics, significantly improving trajectory tracking accuracy. To address this issue, TWO incorporates a trade-off mechanism between tire wear and trajectory tracking accuracy by including an optimization term based on $W_{tw}$ during the model optimization process. This integrated approach effectively balances tire wear and tracking error, providing a practical solution to increased tire wear.
\subsubsection{Case 2}

\begin{table*}[t]
\centering
\setlength{\tabcolsep}{4pt} 
\renewcommand{\arraystretch}{1.5} 
\caption{Comparison of Metrics in Case 2. Best result are in \textbf{bold}, second best are \underline{underlined}. }
\begin{tabular}{c|l|c|cccc|cccc}
\hline
\hline
& \textbf{Method}
& \textbf{$\Omega$}
& \textbf{$W_{tw}$ ($J$)}
& \textbf{$W_{\alpha}$ ($J$)}
& \textbf{$W_{s}$ ($J$)}
& \textbf{$W_{t}$ ($J$)}
& \textbf{$\Bar{e}$}
& \textbf{$e_x$ ($cm$)}
& \textbf{$e_y$ ($cm$)}
& \textbf{$e_{\varphi}$ (°)} \\ \hline 

\multirow{5}{*}{\rotatebox{90}{10 km/h}}
& Kinematic\cite{zhang2016study}
& 7.29 × 10\textsuperscript{5}
& 4.77 × 10\textsuperscript{9}
& \cellcolor[HTML]{D5F5E3} \underline{2.24 × 10\textsuperscript{8}}
& 5.82 × 10\textsuperscript{8}
& 3.96 × 10\textsuperscript{9}
& 111.49
& 189.59
& 92.23
& 52.66 \\

& DH\cite{du2024hierarchical}
& 6.39 × 10\textsuperscript{4}
& 1.99 × 10\textsuperscript{9}
& 4.14 × 10\textsuperscript{8}
& \cellcolor[HTML]{D5F5E3} \underline{1.55 × 10\textsuperscript{9}}
& 2.58 × 10\textsuperscript{7}
& 36.33
& 20.34
& 41.09
& 47.55 \\

& TCT\cite{tang2024all}
& 6.44 × 10\textsuperscript{4}
& 1.94 × 10\textsuperscript{9}
& \cellcolor[HTML]{A9DFBF} \textbf{7.80 × 10\textsuperscript{7}}
& 1.86 × 10\textsuperscript{9}
& 4.54 × 10\textsuperscript{6}
& 44.41
& 57.83
& 34.26
& \cellcolor[HTML]{A9DFBF} \textbf{41.15} \\

& NTWO (Our)
& \cellcolor[HTML]{D5F5E3} \underline{5.96 × 10\textsuperscript{4}}
& \cellcolor[HTML]{D5F5E3} \underline{1.84 × 10\textsuperscript{9}}
& 2.44 × 10\textsuperscript{8}
& 1.57 × 10\textsuperscript{9}
& \cellcolor[HTML]{D5F5E3} \underline{2.55 × 10\textsuperscript{7}}
& \cellcolor[HTML]{A9DFBF} \textbf{27.09}
& \cellcolor[HTML]{A9DFBF} \textbf{9.73}
& \cellcolor[HTML]{A9DFBF} \textbf{27.95}
& \cellcolor[HTML]{D5F5E3} \underline{43.60} \\

& TWO (Our)
& \cellcolor[HTML]{A9DFBF} \textbf{5.71 × 10\textsuperscript{4}}
& \cellcolor[HTML]{A9DFBF} \textbf{1.66 × 10\textsuperscript{9}}
& 3.07 × 10\textsuperscript{8}
& \cellcolor[HTML]{A9DFBF} \textbf{1.33 × 10\textsuperscript{9}}
& \cellcolor[HTML]{A9DFBF} \textbf{2.50 × 10\textsuperscript{7}}
& \cellcolor[HTML]{D5F5E3} \underline{29.23}
& \cellcolor[HTML]{D5F5E3} \underline{12.74}
& \cellcolor[HTML]{D5F5E3} \underline{29.36}
& 45.59 \\ \hline
\end{tabular}
\label{tab:comparison2}
\vspace{3pt}
\end{table*}
\begin{figure}
    \centering
    \includegraphics[width=1\linewidth]{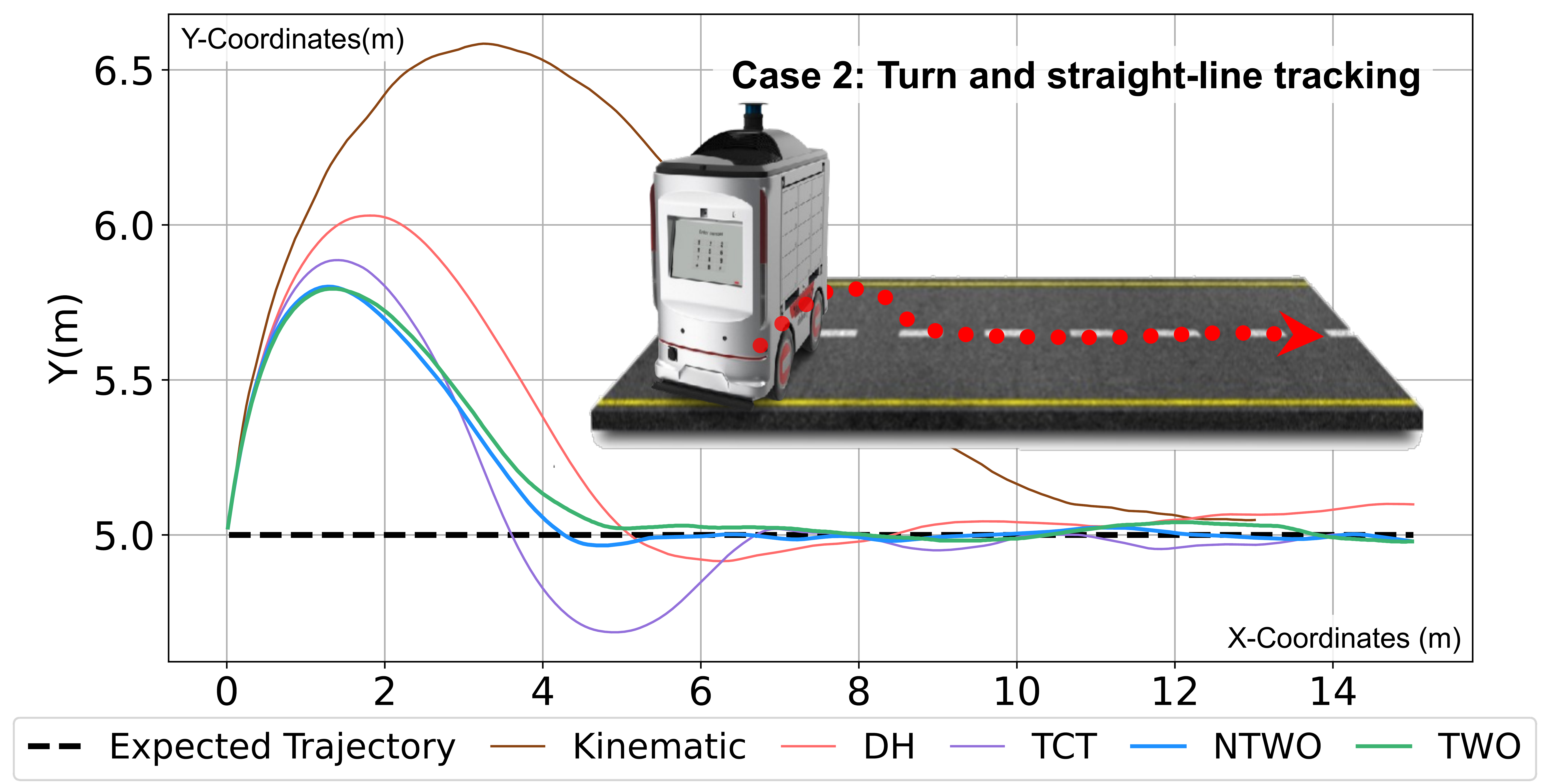}
    \caption{Trajectory Comparison in Case 2}
    \label{fig:TrajectoryComparison2_1}
    \vspace{-15pt}
\end{figure}
\begin{figure}
    \centering
    \includegraphics[width=1\linewidth]{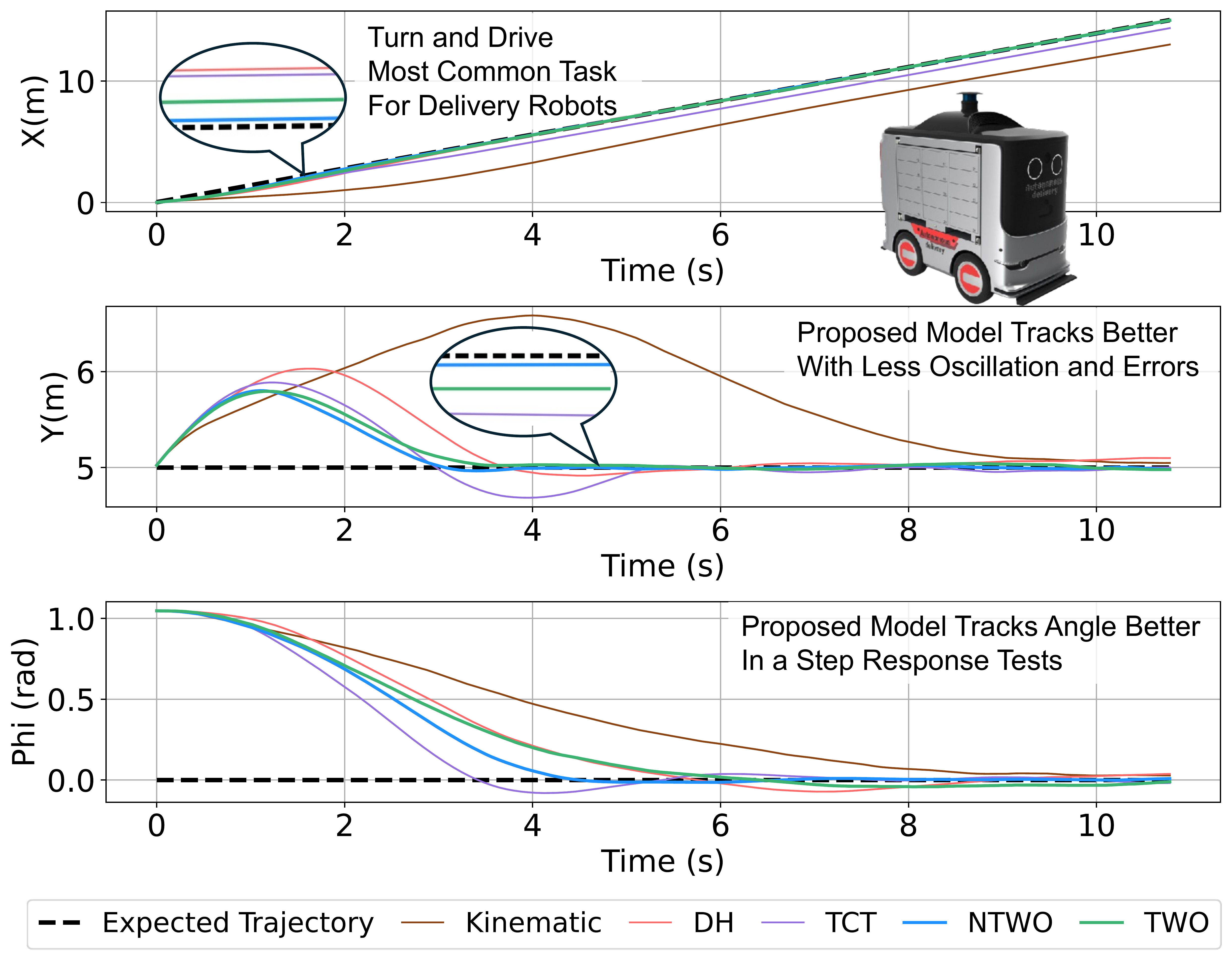}
    \caption{Trajectory Comparison with Time in Case 2}
    \label{fig:TrajectoryComparison2_2}
    \vspace{-15pt}
\end{figure}
In this case, the gross vehicle weight is set to 12 Tons, and we conducted experiments using a linear trajectory at 10 km/h with a control period of (T=0.01s). The initial heading angle was initialized at 60 degrees to evaluate the controller's robustness. The resulting trajectory tracking performance is illustrated in {Fig~\ref{fig:TrajectoryComparison2_1}} and {Fig~\ref{fig:TrajectoryComparison2_2}}, and the total tire wear results are presented in {Table~\ref{tab:comparison2}}.

The results demonstrate that the cumulative tire wear work for the TWO is \(1.66 \times 10^9\) J, representing a 65.20\% reduction compared to the Kinematic Method's \(4.77 \times 10^9\) J, and achieving 16.58\% and 14.43\% reductions compared to the DH and TCT methods, respectively. Moreover, the TWO maintains the optimal Performance Balance Index $\Omega$ at \( 5.71 \times 10^4 \). This demonstrates the TWO's adaptability and superior performance across a wide range of operational scenarios.

\subsubsection{Case 3}
\begin{figure}
    \centering
    \includegraphics[width=1\linewidth]{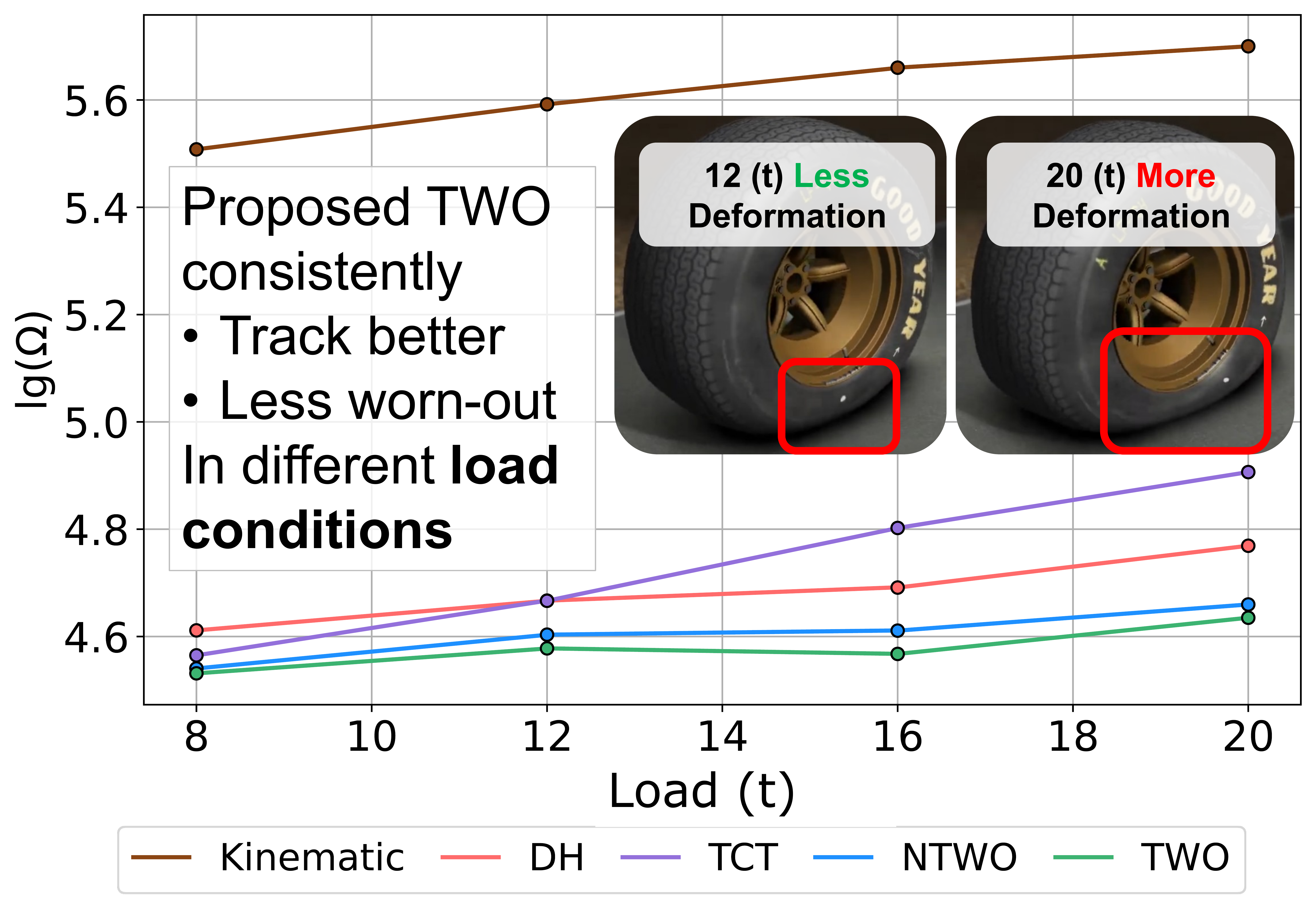}
    \caption{Performance Balance Comparison in Case 3}
    \label{fig:TrajectoryComparison3}
\end{figure}

In this case, the gross vehicle weight varies from 8 Tons to 20 Tons. Experiments were conducted using the same trajectory as in Case 1, at a speed of 30 km/h, with a control period of (T=0.01s). and the relationship between Load and $lg(\Omega)$ is visualized using B-spline interpolation in {Fig~\ref{fig:TrajectoryComparison3}}.

Taking the case with a gross vehicle weight of 12 Tons as an example, the cumulative tire wear work for the TWO is \( 9.33 \times 10^8 \) J, achieving a 19.77\% reduction compared to NTWO's \( 8.80 \times 10^8 \) J. Although the Average RMSE $\Bar{e}$ increases from 4.83 to 11.74, these values remain within acceptable limits. Furthermore, the Performance Balance Index $\Omega$ improves from \( 3.47 \times 10^4 \) to \( 3.40 \times 10^4 \), demonstrating enhanced overall performance.

The results indicate that while the total tire wear work $W_{tw}$ increases with vehicle load, the TWO consistently demonstrates superior overall performance across all load conditions.
\begin{table*}[t]
\centering
\setlength{\tabcolsep}{4pt} 
\renewcommand{\arraystretch}{1.5} 
\caption{Comparison of Metrics in Case 3. Best results are in \textbf{bold}, second best are \underline{underlined}. }
\begin{tabular}{c|l|c|cccc|cccc}
\hline
& \textbf{Method}
& \textbf{$\Omega$}
& \textbf{$W_{tw}$ ($J$)}
& \textbf{$W_{\alpha}$ ($J$)}
& \textbf{$W_{s}$ ($J$)}
& \textbf{$W_{t}$ ($J$)}
& \textbf{$\Bar{e}$}
& \textbf{$e_x$ ($cm$)}
& \textbf{$e_y$ ($cm$)}
& \textbf{$e_{\varphi}$ (°)} \\ \hline 

\multirow{5}{*}{\rotatebox{90}{8 Tons}}
& Kinematic\cite{zhang2016study}
& 3.22 × 10\textsuperscript{5}  
& \cellcolor[HTML]{A9DFBF} \textbf{2.71 × 10\textsuperscript{8}}  
& \cellcolor[HTML]{A9DFBF} \textbf{7.28 × 10\textsuperscript{7}}  
& \cellcolor[HTML]{A9DFBF} \textbf{1.15 × 10\textsuperscript{8}}  
& 8.34 × 10\textsuperscript{7}  
& 382.21  
& 779.11  
& 307.33  
& 60.19  \\

& DH\cite{du2024hierarchical}
& 4.09 × 10\textsuperscript{4}  
& 8.97 × 10\textsuperscript{8}  
& 2.21 × 10\textsuperscript{8}  
& 6.69 × 10\textsuperscript{8}  
& \cellcolor[HTML]{D5F5E3} \underline{7.12 × 10\textsuperscript{6}}  
& 22.43  
& 27.36  
& 28.21  
& 11.72  \\

& TCT\cite{tang2024all}
& 3.67 × 10\textsuperscript{4}  
& 7.57 × 10\textsuperscript{8}  
& 2.25 × 10\textsuperscript{8}  
& \cellcolor[HTML]{D5F5E3} \underline{5.23 × 10\textsuperscript{8}}  
& 9.06 × 10\textsuperscript{6}  
& 17.93  
& 27.47  
& 13.18  
& 13.15  \\

& NTWO (Our)
& \cellcolor[HTML]{D5F5E3} \underline{3.47 × 10\textsuperscript{4}}  
& 8.80 × 10\textsuperscript{8}  
& 1.50 × 10\textsuperscript{8}  
& 7.21 × 10\textsuperscript{8}  
& 8.13 × 10\textsuperscript{6}  
& \cellcolor[HTML]{A9DFBF} \textbf{4.83}  
& \cellcolor[HTML]{A9DFBF} \textbf{4.89}  
& \cellcolor[HTML]{A9DFBF} \textbf{6.07}  
& \cellcolor[HTML]{A9DFBF} \textbf{3.51}  \\

& TWO (Our)
& \cellcolor[HTML]{A9DFBF} \textbf{3.40 × 10\textsuperscript{4}}  
& \cellcolor[HTML]{D5F5E3} \underline{7.06 × 10\textsuperscript{8}}  
& \cellcolor[HTML]{D5F5E3} \underline{1.38 × 10\textsuperscript{8}}  
& 5.61 × 10\textsuperscript{8}  
& \cellcolor[HTML]{A9DFBF} \textbf{7.02 × 10\textsuperscript{6}}  
& \cellcolor[HTML]{D5F5E3} \underline{11.74}  
& \cellcolor[HTML]{D5F5E3} \underline{16.29}  
& \cellcolor[HTML]{D5F5E3} \underline{11.39}  
& \cellcolor[HTML]{D5F5E3} \underline{7.53}  \\
\hline

\multirow{5}{*}{\rotatebox{90}{12 Tons}}
& Kinematic\cite{zhang2016study}
& 3.91 × 10\textsuperscript{5}  
& \cellcolor[HTML]{A9DFBF} \textbf{4.09 × 10\textsuperscript{8}}  
& \cellcolor[HTML]{A9DFBF} \textbf{1.06 × 10\textsuperscript{8}}  
& \cellcolor[HTML]{A9DFBF} \textbf{1.80 × 10\textsuperscript{8}}  
& 1.23 × 10\textsuperscript{8}  
& 373.11  
& 755.26  
& 304.39  
& 59.69  \\

& DH\cite{du2024hierarchical}
& 4.64 × 10\textsuperscript{4}  
& 1.32 × 10\textsuperscript{9}  
& 2.98 × 10\textsuperscript{8}  
& 1.01 × 10\textsuperscript{9}  
& 1.08 × 10\textsuperscript{7}  
& 11.85  
& 17.48  
& 12.64  
& 5.43  \\

& TCT\cite{tang2024all}
& 4.64 × 10\textsuperscript{4}  
& 1.27 × 10\textsuperscript{9}  
& 8.37 × 10\textsuperscript{8}  
& \cellcolor[HTML]{D5F5E3} \underline{4.09 × 10\textsuperscript{8}}  
& 2.88 × 10\textsuperscript{7}  
& 13.77  
& 26.29  
& 13.63  
& \cellcolor[HTML]{A9DFBF} \textbf{1.39}  \\

& NTWO (Our)
& \cellcolor[HTML]{D5F5E3} \underline{4.01 × 10\textsuperscript{4}}  
& 1.15 × 10\textsuperscript{9}  
& 2.06 × 10\textsuperscript{8}  
& 9.39 × 10\textsuperscript{8}  
& \cellcolor[HTML]{D5F5E3} \underline{9.81 × 10\textsuperscript{6}}  
& \cellcolor[HTML]{A9DFBF} \textbf{5.28}  
& \cellcolor[HTML]{A9DFBF} \textbf{6.64}  
& \cellcolor[HTML]{A9DFBF} \textbf{4.43}  
& \cellcolor[HTML]{D5F5E3} \underline{4.78}  \\

& TWO (Our)
& \cellcolor[HTML]{A9DFBF} \textbf{3.78 × 10\textsuperscript{4}}  
& \cellcolor[HTML]{D5F5E3} \underline{9.33 × 10\textsuperscript{8}}  
& \cellcolor[HTML]{D5F5E3} \underline{1.76 × 10\textsuperscript{8}}  
& 7.47 × 10\textsuperscript{8}  
& \cellcolor[HTML]{A9DFBF} \textbf{9.66 × 10\textsuperscript{6}}  
& \cellcolor[HTML]{D5F5E3} \underline{8.51}  
& \cellcolor[HTML]{D5F5E3} \underline{12.84}  
& \cellcolor[HTML]{D5F5E3} \underline{7.70}  
& 4.99  \\
\hline

\multirow{5}{*}{\rotatebox{90}{16 Tons}}
& Kinematic\cite{zhang2016study}
& 4.57 × 10\textsuperscript{5}  
& \cellcolor[HTML]{A9DFBF} \textbf{5.48 × 10\textsuperscript{8}}  
& \cellcolor[HTML]{A9DFBF} \textbf{1.38 × 10\textsuperscript{8}}  
& \cellcolor[HTML]{A9DFBF} \textbf{2.43 × 10\textsuperscript{8}}  
& 1.67 × 10\textsuperscript{8}  
& 381.61  
& 763.45  
& 320.23  
& 61.13  \\

& DH\cite{du2024hierarchical}
& 4.91 × 10\textsuperscript{4}  
& 1.67 × 10\textsuperscript{9}  
& 3.39 × 10\textsuperscript{8}  
& 1.32 × 10\textsuperscript{9}  
& 1.44 × 10\textsuperscript{7}  
& 6.29  
& 7.40  
& 8.39  
& \cellcolor[HTML]{D5F5E3} \underline{3.07}  \\

& TCT\cite{tang2024all}
& 6.34 × 10\textsuperscript{4}  
& 2.35 × 10\textsuperscript{9}  
& 3.80 × 10\textsuperscript{8}  
& 1.96 × 10\textsuperscript{9}  
& \cellcolor[HTML]{A9DFBF} \textbf{1.37 × 10\textsuperscript{7}}  
& 14.61  
& 26.65  
& 13.29  
& 3.88  \\

& NTWO (Our)
& \cellcolor[HTML]{D5F5E3} \underline{4.08 × 10\textsuperscript{4}}  
& 1.58 × 10\textsuperscript{9}  
& 2.65 × 10\textsuperscript{8}  
& 1.30 × 10\textsuperscript{9}  
& 1.40 × 10\textsuperscript{7}  
& \cellcolor[HTML]{A9DFBF} \textbf{1.32}  
& \cellcolor[HTML]{A9DFBF} \textbf{1.65}  
& \cellcolor[HTML]{A9DFBF} \textbf{1.49}  
& \cellcolor[HTML]{A9DFBF} \textbf{0.80}  \\

& TWO (Our)
& \cellcolor[HTML]{A9DFBF} \textbf{3.69 × 10\textsuperscript{4}}  
& \cellcolor[HTML]{D5F5E3} \underline{1.04 × 10\textsuperscript{9}}  
& \cellcolor[HTML]{D5F5E3} \underline{2.27 × 10\textsuperscript{8}}  
& \cellcolor[HTML]{D5F5E3} \underline{8.00 × 10\textsuperscript{8}}  
& \cellcolor[HTML]{D5F5E3} \underline{1.39 × 10\textsuperscript{7}}  
& \cellcolor[HTML]{D5F5E3} \underline{3.88}  
& \cellcolor[HTML]{D5F5E3} \underline{3.93}  
& \cellcolor[HTML]{D5F5E3} \underline{3.44}  
& 4.26  \\
\hline

\multirow{5}{*}{\rotatebox{90}{20 Tons}}
& Kinematic\cite{zhang2016study}
& 5.01 × 10\textsuperscript{5}  
& \cellcolor[HTML]{A9DFBF} \textbf{6.69 × 10\textsuperscript{8}}  
& \cellcolor[HTML]{A9DFBF} \textbf{1.67 × 10\textsuperscript{8}}  
& \cellcolor[HTML]{A9DFBF} \textbf{2.96 × 10\textsuperscript{8}}  
& 2.06 × 10\textsuperscript{8}  
& 375.87  
& 743.45  
& 322.39  
& 61.77  \\

& DH\cite{du2024hierarchical}
& 5.88 × 10\textsuperscript{4}  
& 2.35 × 10\textsuperscript{9}  
& 5.43 × 10\textsuperscript{8}  
& 1.79 × 10\textsuperscript{9}  
& 1.79 × 10\textsuperscript{7}  
& 6.83  
& 9.28  
& 7.45  
& \cellcolor[HTML]{D5F5E3} \underline{3.77}  \\

& TCT\cite{tang2024all}
& 8.06 × 10\textsuperscript{4}  
& 3.79 × 10\textsuperscript{9}  
& 5.37 × 10\textsuperscript{8}  
& 3.23 × 10\textsuperscript{9}  
& \cellcolor[HTML]{A9DFBF} \textbf{1.69 × 10\textsuperscript{7}}  
& 14.94  
& 25.98  
& 14.14  
& 4.69  \\

& NTWO (Our)
& \cellcolor[HTML]{D5F5E3} \underline{4.57 × 10\textsuperscript{4}}  
& 2.10 × 10\textsuperscript{9}  
& 3.56 × 10\textsuperscript{8}  
& 1.72 × 10\textsuperscript{9}  
& \cellcolor[HTML]{D5F5E3} \underline{1.76 × 10\textsuperscript{7}}  
& \cellcolor[HTML]{A9DFBF} \textbf{0.98}  
& \cellcolor[HTML]{A9DFBF} \textbf{1.33}  
& \cellcolor[HTML]{A9DFBF} \textbf{1.05}  
& \cellcolor[HTML]{A9DFBF} \textbf{0.57}  \\

& TWO (Our)
& \cellcolor[HTML]{A9DFBF} \textbf{4.32 × 10\textsuperscript{4}}  
& \cellcolor[HTML]{D5F5E3} \underline{1.33 × 10\textsuperscript{9}}  
& \cellcolor[HTML]{D5F5E3} \underline{3.40 × 10\textsuperscript{8}}  
& \cellcolor[HTML]{D5F5E3} \underline{9.72 × 10\textsuperscript{8}}  
& 1.79 × 10\textsuperscript{7}  
& \cellcolor[HTML]{D5F5E3} \underline{5.40}  
& \cellcolor[HTML]{D5F5E3} \underline{5.63}  
& \cellcolor[HTML]{D5F5E3} \underline{4.40}  
& 6.18  \\
\hline

\end{tabular}
\label{tab:comparison3}
\vspace{3pt}
\end{table*}

\begin{Remark}
To ensure a fair comparison between the five methods (\textbf{Kinematic}, \textbf{DH}, \textbf{TCT}, \textbf{NTWO}, and \textbf{TWO}), the experimental setup and evaluation metrics were carefully designed to eliminate bias. Specifically, the vehicle model, trajectory, initial conditions, and control parameters were uniformly applied across all methods. Additionally, tire wear work calculations and trajectory tracking errors were evaluated using the same equations and assumptions to ensure consistency. By maintaining identical conditions for each method, the results objectively reflect the advantages and trade-offs of the proposed approaches.
\end{Remark}

\section{Conclusion}\label{sec:5}
This paper proposes a tracking model for Multi-Axle Swerve-Drive AMRs based on dynamic analysis, incorporating a tire wear model and tire dynamics model based on the Pacejka Magic Formula. The problem is formulated as  {an} optimization problem and solved using  {a} simulated annealing algorithm.

Through Matlab simulations, we compared the Kinematic Model (Kinematic), Heng Du’s Dynamic Model (DH), Congteng Tang’s Dynamic Model (TCT), Proposed Dynamic Model With Tire Wear Optimization (TWO) and the Proposed Dynamic Model Neglecting Tire Wear Optimization (NTWO). The results demonstrate that The Proposed Dynamic Model With Tire Wear Optimization can effectively reduce tire wear while slightly compromising tracking accuracy. 

Furthermore, the TWO demonstrates superior Performance Balance Index $\Omega$ across various operating conditions, including speeds from 5-35 km/h and gross vehicle weights from 8-20 Tons, demonstrating the robust applicability of our proposed method.

Since the current proposed method assumes uniform load distribution across all tires, it does not account for load variations during wheel motion due to acceleration changes. In future work, we plan to incorporate tire load dynamics into the model to further improve its accuracy.

\section{Acknowledgment}
This research is supported by the National Research Foundation, Singapore, under its Medium-Sized Center for Advanced Robotics Technology Innovation (CARTIN).

\balance
\bibliographystyle{elsarticle-num}
\bibliography{ref_JAI}
\vspace{40pt}
\bio{ htx.jpg }
\textbf{Tianxin Hu} Tianxin Hu received his Bachelor's degree in Vehicle Engineering from Wuhan University of Science and Technology, Wuhan, China, in 2024. He is currently pursuing a Master's degree in Computer Control \& Automation at Nanyang Technological University, Singapore. His research interests include autonomous driving, machine vision, path planning, path tracking, robotic control, and multi-axle vehicle control.
\endbio

\bio{xxh.jpg}
\textbf{Xinhang Xu}  received his Bachelor of Engineering in Automation from South China University of Technology, Guangzhou, China and the M.Sc. degree in electrical and electronic engineering from Nanyang Technological University, Singapore, in 2023. His research interests include modeling and control of aerial and ground robots, motion planning, and multi-agent systems.
\endbio

\bio{tm.jpg}
\textbf{Thien-Minh Nguyen }   is currently a Research Assistant Professor at the Centre for Advanced Robotics Technology Innovation (CARTIN), NTU, Singapore. He was the Wallenberg-NTU Presidential Postdoctoral Fellow at NTU and KTH Royal Institute of Technology from 2020 to 2023. He received the 2020 EEE Best Thesis Award (Innovation Track), and the 2nd Prize at the 2021 Hilti SLAM Challenge (Academic Track). His research interests include autonomous systems, robot perception, navigation, and learning. He is the co-organizer of the CARIC challenge at CDC 2023.
\endbio

\bio{lf.jpg}
\textbf{Fen Liu}   received the M.S. and Ph.D. degrees in
2020 and 2023, respectively, from the School of
Automation, Guangdong University of Technology,
Guangzhou, China. She is currently a research fellow
at the School of Electrical and Electronic Engineering, Nanyang Technological University, Singapore.
Her research interests include robust estimation,
target encirclement, cooperative control, and antisynchronization control.
\endbio

\bio{yuanshenghaibiophoto.png}
\textbf{Yuan Shenghai} is a senior research fellow with the School of EEE, Nanyang Technological University, Singapore. He received his B.S. and Ph.D. degrees in EEE in 2013 and 2019, respectively. He is currently with the Centre for Advanced Robotics Technology Innovation (CARTIN) at Nanyang Technological University.
Dr Yuan`s research focuses on robotics perception, UAV, and navigation. He has contributed over 60 papers in TRO, IJRR, TIE, RAL, ICRA, CVPR, ICCV, NeurIPS, IROS, etc. He currently serves as an associate editor for Unmanned System Journal and guest editor of the Electronics Special issue on Advance Technologies of Navigation for Intelligent Vehicles. He is the organizer for the CARIC Drone Swarm Challenge in CDC 2023 and the UG2 Anti-Drone Challenge in CVPR 2024.
\endbio
\vspace{40pt}
\bio{Lihua_Xie.jpg}
\textbf{Lihua Xie} received the Ph.D. degree in electrical engineering from the University of Newcastle, Australia, in 1992. Since 1992, he has been with the School of Electrical and Electronic Engineering, Nanyang Technological University, Singapore, where he is currently a professor and Director, Delta-NTU Corporate Laboratory for Cyber-Physical Systems and Director, Center for Advanced Robotics Technology Innovation. He served as the Head of the Division of Control and Instrumentation from July 2011 to June 2014. He held teaching appointments in the Department of Automatic Control, Nanjing University of Science and Technology from 1986 to 1989.

Dr Xie’s research interests include robust control and estimation, networked control systems, multi-agent networks, localization and unmanned systems. He is an Editor-in-Chief for Unmanned Systems and has served as Editor of the IET Book Series in Control and Associate Editor of a number of journals, including IEEE Transactions on Automatic Control, Automatica, IEEE Transactions on Control Systems Technology, IEEE Transactions on Network Control Systems, and IEEE Transactions on Circuits and Systems-II. He was an IEEE Distinguished Lecturer (Jan 2012--Dec 2014). Dr Xie is Fellow of Academy of Engineering Singapore, IEEE, IFAC, and CAA.
\endbio

\end{document}